\documentclass[]{antgroup}

\PassOptionsToPackage{numbers, compress}{natbib}
\usepackage{antgroup}

\usepackage{graphicx}
\usepackage{tikz}
\usepackage{todonotes}
\usepackage{multirow}
\usepackage{amsmath}
\usepackage{cleveref}
\usepackage{subcaption}
\usepackage{multicol}
\usepackage{amssymb}
\usepackage{array}
\usepackage{bm}
\usepackage{enumitem}
\usepackage{wrapfig}
\usepackage{algorithmic}
\usepackage{algorithm}
\usepackage{tabularx}
\usepackage{bbm}
\usepackage{makecell}
\usepackage{colortbl}

\usepackage[normalem]{ulem}
\useunder{\uline}{\ul}{}

\usepackage{amsmath,amsfonts,bm}

\def\eqref#1{equation~\ref{#1}}

\def\1{\bm{1}}

\DeclareMathAlphabet{\mathsfit}{\encodingdefault}{\sfdefault}{m}{sl}
\SetMathAlphabet{\mathsfit}{bold}{\encodingdefault}{\sfdefault}{bx}{n}

\newcommand*\justify{%
  \fontdimen2\font=0.4em
  \fontdimen3\font=0.2em
  \fontdimen4\font=0.1em
  \fontdimen7\font=0.1em
  \hyphenchar\font=`\-
}

\renewcommand{\texttt}[1]{%
  \begingroup
  \ttfamily
  \begingroup\lccode`~=`/\lowercase{\endgroup\def~}{/\discretionary{}{}{}}%
  \begingroup\lccode`~=`[\lowercase{\endgroup\def~}{[\discretionary{}{}{}}%
  \begingroup\lccode`~=`.\lowercase{\endgroup\def~}{.\discretionary{}{}{}}%
  \catcode`/=\active\catcode`[=\active\catcode`.=\active
  \justify\scantokens{#1\noexpand}%
  \endgroup
}

\usepackage{amsmath}
\usepackage{amssymb}
\usepackage{amsfonts}                               
\usepackage{amsthm}
\usepackage[mathcal]{eucal}
\usepackage{mathrsfs}
\usepackage{bm}                                     
\usepackage{blkarray}                               
\usepackage{nicefrac}                               

\usepackage{wrapfig}
\usepackage{graphicx}                               
\usepackage{caption}
\usepackage{cleveref}

\usepackage{tikz}                                          
\usepackage{circuitikz}
\usetikzlibrary{patterns}
\usetikzlibrary{positioning,calc,fit,decorations.pathmorphing,shapes.geometric, shapes.gates.logic.US, calc}
\usetikzlibrary{arrows,arrows.meta,decorations.markings,shapes,shapes.arrows}
\usetikzlibrary{decorations,decorations.pathreplacing}
\usetikzlibrary{backgrounds}
\usepackage{pgfplots}
\usepackage{pgfplotstable}
\usepgfplotslibrary{groupplots}
\usepackage{scalefnt}
\pgfplotsset{compat=newest}
\usepackage{xcolor}
\definecolor{firstcolor}{HTML}{C3423F}
\definecolor{secondcolor}{HTML}{2A4B8C}

\title{Inclusion Arena: An Open Platform for Evaluating Large Foundation Models with Real-World Apps}

\author{Kangyu Wang$^{1,2*}$, Hongliang He$^{1,3,4*}$, Lin Liu$^{1*}$, Ruiqi Liang$^{1}$, Zhenzhong Lan$^{1,4\dag}$, Jianguo Li$^{1\dag}$}
\affiliation{$^1$Inclusion AI\quad}
\affiliation{$^2$Shanghai Jiao Tong University\quad}
\affiliation{$^3$Zhejiang University\quad}
\affiliation{$^4$Westlake University}
\footnotetext{$^*$Equal Contribution.}
\footnotetext{$^\dag$Corresponding Authors.}
\begin{document}
\maketitle

\begin{abstract}
Large Language Models (LLMs) and Multimodal Large Language Models (MLLMs) have ushered in a new era of AI capabilities, demonstrating near-human-level performance across diverse scenarios. 
While numerous benchmarks (e.g., MMLU) and leaderboards (e.g., Chatbot Arena) have been proposed to help evolve the development of LLMs and MLLMs, most rely on static datasets or crowdsourced general-domain prompts, often falling short of reflecting performance in real-world applications. To bridge this critical gap, we present Inclusion Arena, \textbf{a live leaderboard that ranks models based on human feedback collected directly from AI-powered applications}. Our platform integrates pairwise model comparisons into natural user interactions, ensuring evaluations reflect practical usage scenarios. For robust model ranking, we employ the Bradley-Terry model augmented with two key innovations: (1) Placement Matches, a cold-start mechanism to quickly estimate initial ratings for newly integrated models, and (2) Proximity Sampling, an intelligent comparison strategy that prioritizes battles between models of similar capabilities to maximize information gain and enhance rating stability.
Extensive empirical analyses and simulations demonstrate that Inclusion Arena yields reliable and stable rankings, exhibits higher data transitivity compared to general crowdsourced datasets, and significantly mitigates the risk of malicious manipulation. By fostering an open alliance between foundation models and real-world applications, Inclusion Arena aims to accelerate the development of LLMs and MLLMs truly optimized for practical, user-centric deployments. The platform is publicly accessible at \url{https://www.tbox.cn/about/model-ranking}.
\end{abstract}

\begin{figure}[t]
    \centering
    \includegraphics[width=0.99\linewidth]{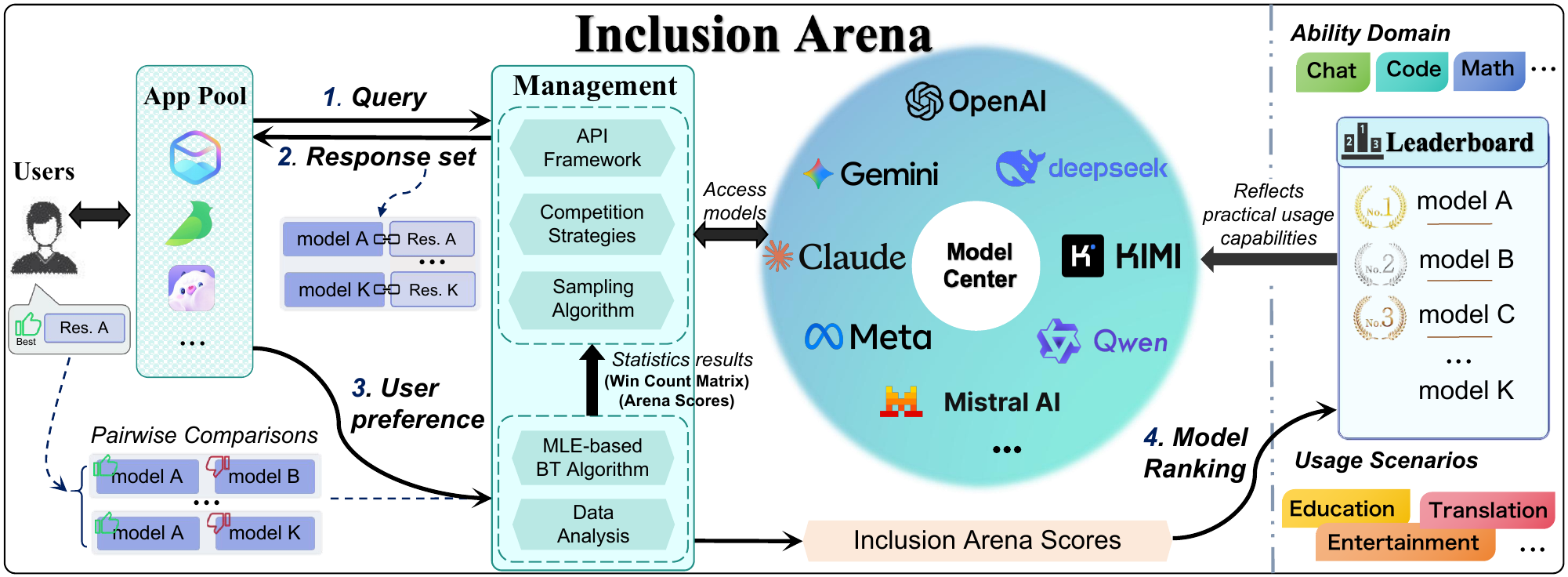}
    \caption{
    Inclusion Arena Pipeline. As a user interacts with an AI application, the system samples K different models in real-time using a proximity sampling algorithm to generate a set of responses for the user's current query. This response set is then presented to the user, who is asked to make a preference judgment. The user's preference is converted into pairwise comparisons and processed by the MLE-based BT algorithm to calculate the Inclusion Arena score for each model. Ultimately, these scores are used to produce a final leaderboard, which can be further broken down into sub-rankings by ability domain and usage scenario, accurately reflecting the models' practical performance in real-world applications.
    }\label{fig:tbox_pipeline}
\end{figure}

\section{Introduction}

In recent years, Large Language Models (LLMs) and Multimodal Large Language Models (MLLMs) have demonstrated human-parity interaction capabilities, reaching a pivotal technical milestone that enables the widespread deployment of AI-powered applications~\citep{openai2023gpt4,team2023gemini,guo2025deepseek,anthropic2023claude,yang2025qwen3,sun2024hunyuan}. This rapid advancement has created an urgent need for systematic evaluation and ranking frameworks to quantify performance differences and facilitate the practical adoption of these models. 

Current evaluation suites and leaderboards can be categorized in several dimensions. 
First, their scope varies across: generic (e.g., OpenLLM~\citep{OSQ}), domain-specific (e.g., LiveCodeBench~\citep{jain2024livecodebench} for coding), or multi-domain (e.g., OpenCompass~\citep{OpenCompass}). Second, the questions used for evaluation could be either static or live. For instance, MMLU~\citep{hend2020mmlu} is a leaderboard with static pre-defined question sets for LLM evaluation, while LiveBench~\citep{white2024livebench} regularly updates questions in a certain period (e.g., weekly or monthly). In contrast to these benchmarks with predefined ground-truth references, Chatbot Arena~\citep{chatbot} introduces a new evaluation paradigm based on human preferences. This approach involves dynamically collecting human preferences over LLM responses to open-ended questions and deriving model rankings from these pairwise comparisons, thereby making the evaluation results more aligned with the real-world usage of LLMs. 

However, live leaderboards like Chatbot Arena face several critical limitations: (1) Their data collection through crowdsourcing platforms primarily captures general-domain interactions, which poorly represent the distribution and complexity of real-world application scenarios. (2) The current battle sampling methodology exhibits significant imbalance, as shown in~\citet{singh2025leaderboardillusion}, some models receive insufficient comparisons, leading to biased rating estimates. (3) Unrestricted prompt submission on the open platform creates potential vulnerabilities for data manipulation, as each question-answer pair becomes a separate data point.

To address these gaps, we propose Inclusion Arena, a live leaderboard that bridges real-world AI-powered applications with state-of-the-art LLMs and MLLMs. Unlike crowdsourced platforms, our system randomly triggers model battles during multi-turn human-AI dialogues in real-world apps. We collect human feedback on LLM responses and, following Chatbot Arena~\citep{chatbot}, employ the Bradley-Terry model~\citep{bradley1952rank} to compute model rankings. The user registration process for apps and the randomized triggering mechanism for model battles in multi-turn dialogues increase the difficulty of data manipulation, thereby enhancing the reliability of data from Inclusion Arena. 
Furthermore, to improve data effectiveness, we introduce a novel model sampling strategy called \textbf{Proximity Sampling}, which prioritizes comparisons between models with similar rating scores. This approach allocates limited resources to model comparisons with higher uncertainty, maximizing the information gained from each pairwise comparison. During this process, we also assign higher sampling weights to models with fewer comparisons to ensure fair sampling and data balance. Additionally, we design a cold-start process based on \textbf{Placement Matches} for newly registered models to quickly estimate their approximate rating levels, enabling rapid integration into our platform. We present the complete pipeline in Figure \ref{fig:tbox_pipeline}.

The number of initially integrated AI-powered applications is limited, but we aim to build an open alliance to expand the ecosystem. With more apps registered, we can collect millions of queries and human feedback daily, enabling stable and usage-oriented model rankings. Currently, our platform has integrated two apps, with over 46,611 active users. Our platform also includes more than 49 models and has gathered over a million model battles. Over time, we plan to integrate more apps and models, reaching a broader user base. In this paper, we selected data up to July 2025, and after collection and filtering, obtained 501,003 pairwise comparisons for model ranking and analysis.
We believe that the Inclusion Arena leaderboard could help 
push the evolution of LLMs/MLLMs in real-world applications. Our contributions can be summarized as follows:
\begin{itemize}[leftmargin=2em]
    \item We propose Inclusion Arena, \textbf{a live leaderboard that ranks models based on real-world user preferences from AI-powered apps}. We present the architecture and engineering details of the platform system.
    \item We introduce a novel and efficient proximity sampling strategy to obtain more informative pairwise comparisons. Additionally, we design a cold-start mechanism, termed Placement Matches, to quickly estimate the approximate ranking of newly integrated models. We further validate the rationale behind our approach through simulation experiments.
    \item We plan to open-source a portion of the user preference data collected on the Inclusion Arena Platform to foster the evolution of AI-powered Apps and facilitate research on improving user experience.
\end{itemize}
\section{Preliminaries}
Platforms such as Chatbot Arena~\citep{chatbot} employ pairwise comparisons (a.k.a. battles), where LLMs are dynamically selected to provide to provide responses to conversational tasks and having users determine the results of the competition. This approach generates rich pairwise comparison data, requiring robust statistical methods to aggregate outcomes into a coherent ranking.

\paragraph{Elo Rating System}
The Elo rating system, originally designed for chess by Arpad Elo~\citep{elo1978rating}, is a dynamic method to quantify the relative skill levels of players in competitive games. Each player maintains a rating score $u$, updated based on pairwise comparison outcomes. The probability that player $i$ defeats player $j$ is modeled as:
\begin{equation}\label{equ:elo}
    P(i  \text{ beats } j) = \frac{1}{1+10^{-(u_i - u_j)/\lambda}},
\end{equation} 
where \( \lambda \) is a scaling factor (e.g., \( \lambda=400 \) in classical Elo). After observing a binary outcome \( S_{ij} \in \{0,1\} \), Elo adjusts the ratings via:
\begin{equation}
    u_i \leftarrow u_i + K \cdot (S_{ij} - P(i \text{ beats } j)), 
\end{equation}
where $K$ is a step-size hyperparameter. Elo’s simplicity and interpretability make it widely adopted, but it relies on heuristic updates and assumes independent pairwise comparisons.

\paragraph{Bradley-Terry (BT) Model}
The Bradley-Terry model~\citep{bradley1952rank} is also a probabilistic framework to rank players from pairwise comparison data. Given $n$ players with strengths $\{ \pi_i = e^{\alpha u_i}\}_{i=1}^n$, the probability that $i$ beats $j$ is:
\begin{equation}\label{equ:BT}
    P(i \text{ beats } j) = \frac{\pi_i}{\pi_i + \pi_j} = \frac{1}{1 + e^{-\alpha(u_i-u_j)}}.
\end{equation} 
Unlike Elo, BT parameters are typically estimated offline via maximum likelihood estimation (MLE), which produces significantly more stable ratings, as mentioned in Chatbot Arena~\citep{chatbot}. Let \( H_{ij} \in \{0, 1\} \) denote the outcome of a pairwise comparison between players \( i \) and \( j \), where \( H_{ij} = 1 \) indicates that \( i \) beats \( j \). The rating scores \( \mathbf{u} = (u_1, \dots, u_n) \) are estimated by: 
\begin{equation}
    \mathbf{u} = \mathop{\arg\min}_{\mathbf{u}} \sum_{i,j} \mathcal{L}\left( H_{ij}, \frac{1}{1 + e^{-\alpha(u_i - u_j)}} \right),  
\end{equation}
where \( \mathcal{L}(\cdot, \cdot) \) is the binary cross entropy. When $\alpha = \ln10/\lambda$, the functional form of the BT probability model becomes identical to that of the Elo system.

It should be emphasized that Elo updates are inherently sequential and path-dependent, producing inconsistent results across different data orderings. In contrast, the BT approach generates unique, order-invariant ratings through global optimization, leading to more stable and statistically consistent parameter estimation.

\paragraph{Disc Decomposition}
While the Elo and Bradley-Terry models are effective for games with strong additive transitive structures, they can struggle to accurately represent games with non-additive transitivity or inherent cyclic relationships (e.g., rock-paper-scissors dynamics). The Disc Decomposition algorithm~\citep{bertrand2023limitationselorealworldgames} offers a more expressive framework. It models the probability of player $i$ beating player $j$ using two-dimensional scores $(u_i, v_i)$ for each player: 
\begin{equation}\label{eq:disc_decom}
P(i \text{ beats } j) = \sigma(u_i v_j - v_i u_j),
\end{equation}
where $\sigma$ is the sigmoid function.  This approach provides richer features, enabling the modeling of both transitive games and cyclic disc games, which may offer more accurate modeling complex battles like those among LLMs.  Visualizations of the learned $(u, v)$ reveal a clear transition: for Elo-like (highly transitive) games, the $(u_i, v_i)$ pairs align near-vertically $(v_i \approx 1)$, making $u_i$ the dominant axis. For cyclic disc games, the $(u_i, v_i)$ pairs are spread out, reflecting the essential two-dimensional nature.

\section{Related Work}
\subsection{Static Benchmarks and Live Leaderboards}
Foundation model evaluation paradigms have evolved from traditional static benchmarks towards dynamic, human-in-the-loop systems. Traditional evaluation of LLMs is based on human-labeled static benchmarks, which assess models on fixed datasets with predefined ground-truth answers. These benchmarks are crucial for systematically comparing model performance across specific tasks or domains:
\begin{itemize}[leftmargin=2em]
    \item Domain-Specific Benchmarks: These benchmarks focus on evaluating a model's proficiency in a particular capability or a narrow task domain. For LLMs, examples include HumanEval~\citep{HumanEval} for code generation, GSM8K~\citep{GSM8K} for mathematical reasoning, and MedBench~\citep{medbench} for medical knowledge. For Vision-Language Models (VLMs), benchmarks include Visual Genome~\citep{visualgenome} for visual grounding, DocVQA~\citep{docvqa} for infographic understanding, MathVision~\citep{mathvision} and MathVista~\citep{MathVista} for multimodal reasoning, and Mind2Web~\citep{mind2web}, WebCanvas~\citep{webcanvas} and WebVoyager~\citep{webvoyager} for GUI-agent, etc.
    \item Multi-Domain and General Benchmarks: These integrate tasks from various areas, aiming for broader coverage. Noteworthy examples for LLMs include MMLU (Massive Multitask Language Understanding)~\citep{hend2020mmlu}, which evaluates knowledge across 57 subjects, and the OpenLLM Leaderboard~\citep{openllm}, which aggregates performance across several open-source datasets covering instruction following, reasoning, and world knowledge. OpenCompass~\citep{OpenCompass} and FlagEval~\citep{flageval} also categorize and evaluate LLM capabilities across multiple dimensions.
\end{itemize}

Most static benchmarks misalign with real-world, open-ended usage due to their limited scope and reliance on synthetic data. To mitigate these limitations and better align evaluations with human perception and real-world utility, live leaderboards with dynamic, human-centric evaluations have emerged. Chatbot Arena~\citep{chatbot}, developed by LMSYS, pioneered this approach, employing crowdsourced pairwise comparisons and Elo ratings to rank models based on user preferences. This approach offers valuable insights into model performance on complex, open-ended questions. Currently, arena-based leaderboards have gained significant attention, with variants like WebDev Arena, RepoChat Arena, Copilot Arena, Agent Arena, and Multimodal Arena extending to specialized fields. Moreover, many leaderboards have launched their own arena versions, including OpenCompass~\citep{OpenCompass}, FlagEval~\citep{flageval}, AGI-Eval~\citep{agieval}, and Artificial Analysis Video Arena~\citep{videoarena}.

\subsection{Large Model-Based Internet Apps}
The widespread adoption of large model-based internet applications provides a crucial source of human interaction and implicit feedback. These applications leverage foundation models to offer diverse functionalities to end-users, generating real-world usage data that is insightful for understanding user preferences and model utility. General-purpose applications, which primarily manifested as chatbots (e.g., ChatGPT~\citep{openai2023gpt4}), serve as conversational AI assistants designed to address a vast array of user queries across various domains. Domain-Specific Applications, which focus on specialized tasks, include AI writing assistants (e.g., Grammarly~\citep{grammarly}, Nova~\citep{nova}), code generation environments (e.g. GitHub Copilot~\citep{githubcopilot}), etc. 

\subsection{Ranking Methods}
Diverse methodologies are employed to translate evaluation data into meaningful model rankings:
For static benchmarks, direct metrics like accuracy, F1-score, BLEU~\citep{bleu}, or pass@k quantify performance against ground truth.
For human feedback and pairwise comparisons, probabilistic ranking models are common:
Bradley-Terry (BT) Model~\citep{bradley1952rank}: This model assigns a latent skill parameter to each model, using a logistic function to estimate win probabilities from pairwise comparisons.
Elo Rating System~\citep{elo1978rating}: Adapted from chess, Elo dynamically updates model ratings based on pairwise comparison outcomes, effectively reflecting relative performance in real time.
TrueSkill~\citep{trueskill2005}: A Bayesian extension of Elo/BT, TrueSkill estimates skill while also quantifying uncertainty, useful for sparse data or team evaluations.
The chosen ranking method significantly influences the stability, interpretability, and fairness of the resulting leaderboards, especially in dynamic, human-in-the-loop settings.

\begin{figure}[!tb]
    \centering
    \includegraphics[width=0.9\textwidth]{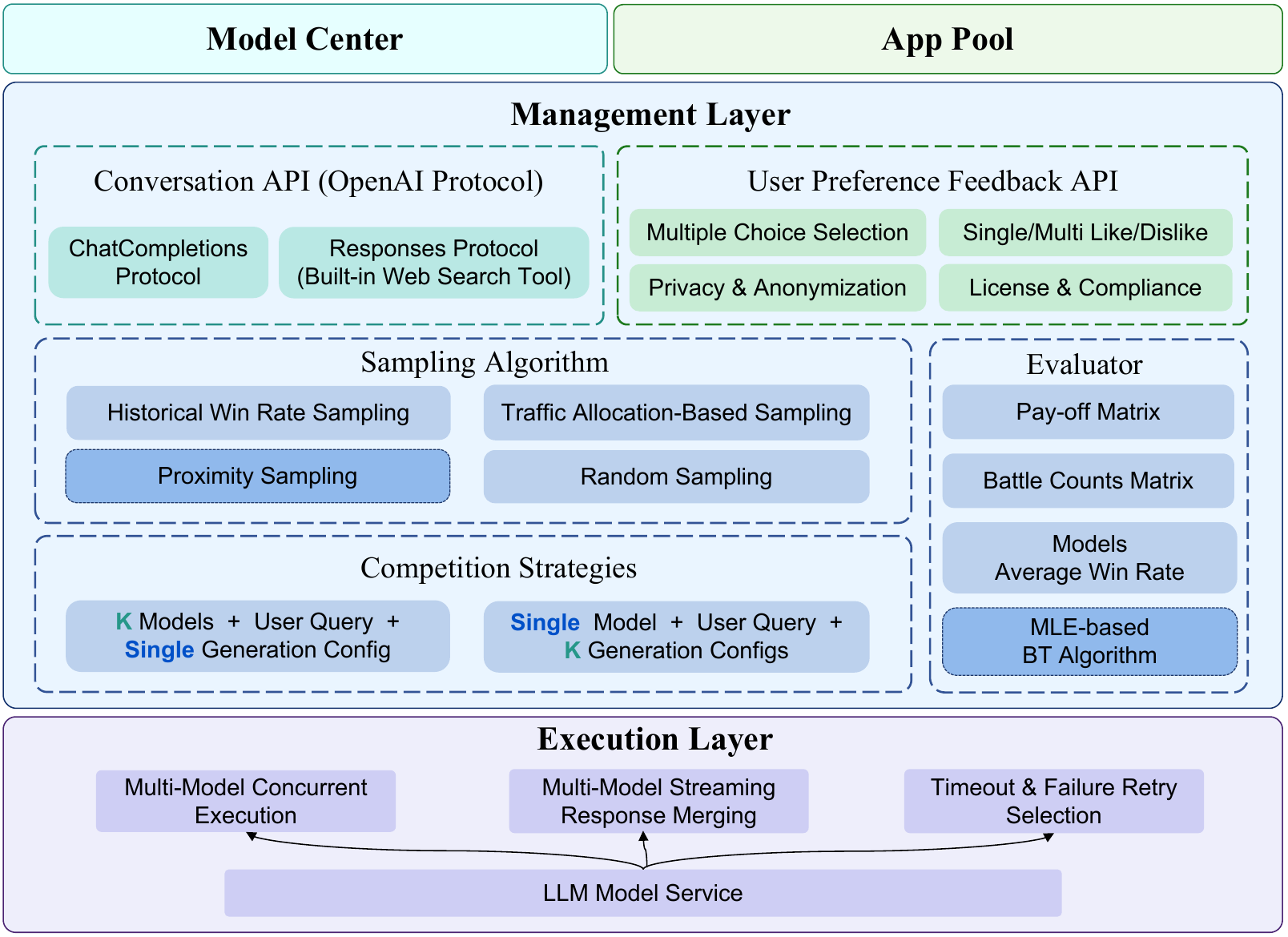}
    \caption{Inclusion Arena Platform Architecture.}
    \label{arch}
\end{figure}

\section{Inclusion Arena Platform: System Architecture}
To effectively capture user preferences for LLMs in real-world applications, we introduce the \textbf{Inclusion Arena Platform}, a system designed to collect preferences and evaluate model performance through natural user interactions. This app-integrated framework improves the fidelity of user feedback and enhances the robustness and applicability of model evaluation and ranking.

\subsection{System Architecture Overview}
The Inclusion Arena Platform integrates real-world user feedback to support comprehensive and scalable evaluation of large-scale AI models. As illustrated in Figure~\ref{arch}, the system is structured into two primary layers: the \textbf{Management Layer} and the \textbf{Execution Layer}, along with two external components: the \textbf{Model Center} and the \textbf{App Pool}. Each component is designed to ensure dynamic interaction, reliable data flow, and meaningful performance analysis within a unified evaluation framework.

\subsection{App Pool and Model Center}
The \textbf{App Pool} serves as the front-end interface where users register, interact with AI-powered applications, and implicitly or explicitly provide feedback on model outputs. These applications span diverse domains and are instrumented with standardized APIs to connect with the backend. As users interact with models through these applications, two key data flows are triggered: (1) Conversational Requests, which encapsulate the user's current query, specific system prompts, and the historical conversation context, are sent to selected models via the Management Layer. (2) User Preference Feedback is collected in real-time during or after interactions. This tight coupling between user behavior and evaluation ensures that feedback signals are grounded in authentic usage scenarios.

The \textbf{Model Center} maintains and orchestrates the lifecycle of all AI models integrated into the platform. It supports both built-in models and externally hosted services via OpenAI-standardized protocols, ensuring interoperability and secure access.
Models are invoked through standardized Conversation APIs, enabling consistent handling across varied deployments. The Model Center provides: (1) Secure access control via API keys; (2) Status management (activation, deactivation, blacklisting/whitelisting); and (3) Integration with internal or third-party hosted LLMs. This modular design enables independent business channels to register and evaluate their own models within a unified infrastructure.

\subsection{Management Layer}
The Management Layer forms the core of the Inclusion Arena Platform, mediating between user interactions and model evaluation. It is composed of four tightly integrated submodules:
\begin{itemize}[leftmargin=2em]
    \item \textbf{Conversation API and User Preference Feedback API:} The Conversation API handles model querying via the ChatCompletions Protocol and the Responses Protocol. The User Preference Feedback API standardizes the collection of user preferences through: Single Like/Dislike, Multiple Likes/Dislikes, \textbf{Multiple Choice Selection} (default setting). These diverse feedback formats are subsequently normalized into a standardized set of pairwise comparisons (e.g., Model A is preferred over Model B) for the evaluator. Collectively, these APIs ensure consistent and structured communication between the App Pool and the backend modules.  \textbf{To ensure ethical data usage}, the feedback pipeline integrates a Privacy \& Anonymization module, which strips personally identifiable information, and a License \& Compliance module that enforces compliance with license agreements and explicit user consent requirements.
    \item \textbf{Sampling Algorithms:} To determine which models or configurations participate in a given interaction, the platform provides a range of sampling strategies, including: Historical Win Rate Sampling, Traffic Allocation-Based Sampling, Random Sampling, and \textbf{Proximity Sampling} (the default, detailed in Section \ref{sec:proximity_sampling}).
    \item \textbf{Competition Strategies.} Inclusion Arena supports two main competition paradigms: $K$ Models + User Query + Single Generation Config; or Single Model + User Query + $K$ Generation Configs. These flexible configurations allow for comparative evaluation of either different models or different prompts/settings for the same model.
    \item \textbf{Evaluator:} The Evaluator module processes user feedback and maintains detailed performance metrics.  It calculates the primary evaluation metric -- Arena Score -- using both Maximum Likelihood Estimation (MLE) and the Bradley-Terry (BT) model. Additionally, it tracks key auxiliary statistics including: payoff matrices, battle count matrices, and average win rates. Crucially, these computed metrics serve as input parameters for the Sampling Algorithms module, forming a feedback loop that enables adaptive model selection in subsequent interactions. To ensure data reliability, the module also incorporates quality filters, which ensure that the rankings reflect the effectiveness of the model and the satisfaction of the user.

\end{itemize}

\subsection{Execution Layer}
The Execution Layer is responsible for dispatching and managing model inference requests. It supports: (1) multi-model concurrent execution; (2) multi-model streaming response merging; and (3) timeout and failure retry selection. 
This layer ensures reliability and responsiveness during real-time interactions, even under varying latency or failure conditions. It acts as the infrastructure backbone for scalable and fault-tolerant model evaluations.

\section{Efficient Model Ranking}

 The Bradley-Terry model provides a robust framework for inferring latent abilities from pairwise comparison outcomes. However, in practical scenarios, particularly with a large and growing number of models, the prospect of exhaustive pairwise comparisons becomes computationally prohibitive and resource-intensive. This highlights a critical need for intelligent battle strategies that maximize information gain within a limited budget.

\subsection{Placement Matches and Proximity Sampling}\label{sec:proximity_sampling}
To enhance the efficiency of the ranking process and maximize the informativeness of each pairwise comparison, we introduce two key components: the \textbf{placement match mechanism} and \textbf{proximity sampling}. Specifically, the placement match mechanism estimates an initial ranking and score for newly registered models, while the proximity sampling constrains pairwise comparisons to models whose estimated ratings fall within a predefined trust region.

\paragraph{Placement Matches} 
We employ placement matches to rapidly estimate an initial rating score for new models through limited comparisons and the binary search algorithm. This process mainly involves the following steps: 
\begin{enumerate}[label=(\arabic*), leftmargin=2em]
    \item Given a set of \( \zeta \) pre-ranked models \( \{M_1, \dots, M_{\zeta}\} \), an initial rating interval \( [l, h] \) ($1\leq l< h \leq \zeta$) is assigned to the new model \( M_{new} \). This interval can either span the entire range of existing models (i.e., \([1, \zeta]\)) or be determined based on available metadata.
    \item A \textbf{binary search} procedure is employed to estimate the rating of \( M_{new} \). In each round, the median model \( M_{mid} \), which is ranked at the midpoint of the current interval \( [M_l, M_h] \), is selected. Platform traffic is then preferentially allocated to conduct \( T \) battles between \( M_{new} \) and \( M_{mid} \). 
    \item 
    Based on the outcome of the comparisons, the rating interval is updated to either \( [M_l, M_{mid}] \) or \( [M_{mid}, M_h] \). The procedure has two termination conditions: it can stop early if the win rate between \( M_{new} \) and \( M_{mid} \) approaches 50\%, or it terminates when fewer than 3 models remain in the interval. Once the procedure terminates, the final rating score \( u_{new} \) for \( M_{new} \) is determined using either Eq. \ref{equ:elo} or \ref{equ:BT}, incorporating the rating score \( u_{mid} \) from the last comparison.
\end{enumerate}

\begin{algorithm}
\caption{Proximity Sampling Algorithm}
\label{alg:proximity_sampling}

\begin{algorithmic}[1]
\REQUIRE Current models $\mathcal{M}=\{M_1,...,M_r\}$ (sorted by score), score vector $\mathbf{u}=[u_1,...,u_r]^T$, battle count matrix $\mathbf{N}$, confidence threshold $h$, temperature $\tau$, desired sample size $K$, minimum proximity model number $n_m$
\ENSURE Sampled model set $\boldsymbol{\phi_s}$

\STATE \textbf{Initialization:} Model sampling weight vector $\mathbf{w} = [w_1, ..., w_r]^T$

\STATE \textcolor{gray}{\# Compute initial sampling weights:}
\FOR{$i = 1$ \TO $r$}
    \STATE $\delta_{M_i} \gets \begin{cases} 
\{ M_j \in \mathcal{M} \mid |u_j - u_i| < h \} & \text{if } |\{ M_j \in \mathcal{M} \mid |u_j - u_i| < h \}| \geq n_m, \\
\{ (n_m -1) \text{ models closest } \text{ to } M_i \} \cup \{ M_i \} & \text{otherwise.}
\end{cases}$ 
    \STATE $n_{\min} \gets \min_{M_j \in \delta_{M_i}} \mathbf{N}_{ij}$
    \STATE $S_{\max} \gets \max(\mathbf{N})$
    \STATE $w_i \gets 1 - n_{\min}/S_{\max}$
\ENDFOR

\STATE Sample initial model $M_t$ according to weights $\mathbf{w}$
\STATE $\boldsymbol{\phi_s} \gets \{M_t\}$
\STATE $\boldsymbol{\phi_r} \gets \delta_{M_t} \setminus \{M_t\}$ \quad  \textcolor{gray}{\# Remaining candidate set}

\STATE
\STATE \textcolor{gray}{\# Iterative sampling of additional models:}

\WHILE{$|\boldsymbol{\phi_s}| < K$ \AND $\boldsymbol{\phi_r} \neq \emptyset$}
    \STATE \textbf{Compute sampling probabilities:}
    \FOR{$M_j \in \boldsymbol{\phi_r}$}
        \STATE $m_j \gets \min_{M_i \in \boldsymbol{\phi_s}} \mathbf{N}_{ji}$ \quad  \textcolor{gray}{\# Min comparison count with selected models}
        \STATE $p_j \gets -m_j/\tau$
    \ENDFOR
    \STATE $\mathbf{p} \gets softmax(\mathbf{p})$ \quad  \textcolor{gray}{\# Normalize $\mathbf{p}$ to probabilities using softmax}
    
    \STATE Sample $M_{new}$ from $\boldsymbol{\phi_r}$ according to probabilities $\mathbf{p}$
    \STATE $\boldsymbol{\phi_s} \gets \boldsymbol{\phi_s} \cup \{M_{new}\}$

    \STATE
    \STATE \textbf{Update remaining candidate set:}
    \STATE $\boldsymbol{\phi_r} \gets \boldsymbol{\phi_r} \setminus \{M_{new}\}$
    \FOR{$M_j \in \boldsymbol{\phi_r}$}
        \IF{$\exists M_i \in \boldsymbol{\phi_s}$ such that $|u_j - u_i| \geq h$}
            \STATE $\boldsymbol{\phi_r} \gets \boldsymbol{\phi_r} \setminus \{M_j\}$
        \ENDIF
    \ENDFOR
\ENDWHILE

\RETURN $\boldsymbol{\phi_s}$

\end{algorithmic}

\end{algorithm}

\paragraph{Proximity Sampling}

We posit that finite comparison resources should be preferentially allocated to acquiring the most informative outcomes, i.e., \textbf{comparisons between models of comparable ability}. 
After the placement matches, we obtain the rating score \( u \) for \( M_{new} \). We then set a trust region threshold \( h \), such that the model selected for battle with \( M_{new} \) should have a rating score \( u' \) satisfying \( |u' - u| < h \).  
This proximity sampling strategy improves the convergence of the Bradley-Terry (BT) system and better distinguishes fine-grained model abilities. When two models possess similar abilities, the outcome of their direct comparison exhibits maximum uncertainty. According to Shannon's definition of entropy, for a binary event with probability \(p\), the entropy \(H = -p \log_2(p) - (1-p) \log_2(1-p)\) is maximized when \(p=0.5\). Conversely, when there is a significant disparity in model capabilities, the outcome is largely deterministic, thus yielding minimal information gain.

In the daily operation of the Inclusion Arena Platform, Proximity Sampling also considers issues such as \textbf{sampling efficiency and data balance}. Our framework extends model comparisons beyond pairwise evaluations (typically involving 3 - 5 model comparisons). Furthermore, we prioritize under-sampled proximity regions by adaptively increasing their selection probabilities.  


Given a sorted set of models \(\mathcal{M} = \{M_1, ..., M_r\}\), a battle count matrix \(\mathbf{N}\) (where \(\mathbf{N}_{ij}\) denotes the number of comparisons between \(M_i\) and \(M_j\)), a sampling size \(K\), and a proximity threshold \(h\), we define the \textbf{proximity interval} \(\delta_{M_i}\) for each model \(M_i \in \mathcal{M}\) as the set of models within a certain neighborhood of \(M_i\). The algorithm proceeds as follows:

\begin{itemize}[leftmargin=2em]
    \item Step 1: Initial Model Sampling.
        \begin{itemize}
            \item  For each model \(M_i \in \mathcal{M}\), considering the Bucket Effect (i.e., to ensure sampling uniformity by prioritizing less-compared pairs), we compute its \textbf{Proximity Comparison Count (PCC)}:  \(n_{i}^{PCC} = \min_{M_j \in \delta_{M_i}} \mathbf{N}_{ij},\)
            which represents the minimum number of comparisons between \(M_i\) and any model in its proximity interval.  
            \item Assign sampling weights to each model inversely proportional to \(n_{i}^{PCC}\) (models with lower PCC have higher weights). 
            \item Sample the initial model \(M_t\) from \(\mathcal{M}\) according to these weights.  
        \end{itemize}
    \item Step 2: Sample additional models.
        \begin{itemize}
            \item  Initialize: Selected model set \(\boldsymbol{\phi_s} = \{M_t\}\) and remaining model set \(\boldsymbol{\phi_r} = \delta_{M_t} \setminus \{M_t\}\).  
            \item Compute Sampling Probabilities for each candidate model \(M_j \in \boldsymbol{\phi_r}\). We compute its 
            \textbf{Effective Comparison Count (ECC)} as: \( n_j^{ECC} =  \min_{M_i \in \boldsymbol{\phi_s}} \mathbf{N}_{ji}, \)  i.e., the minimum comparison count between \(M_j\) and any already-selected model in \(\boldsymbol{\phi_s}\). We assign sampling probabilities inversely proportional to this value.  
            \item Sample \(M_j\) from \(\boldsymbol{\phi_r}\) according to the computed probabilities. Update: \(\boldsymbol{\phi_s} \leftarrow \boldsymbol{\phi_s} \cup \{M_j\}\), \(\boldsymbol{\phi_r} \leftarrow \boldsymbol{\phi_r} \setminus \{M_j\}\).  Then remove any \(M_j \in \boldsymbol{\phi_r}\) that violates the proximity constraint with any \(M_i \in \boldsymbol{\phi_s}\).
        \end{itemize}
    \item Step 3: Repeat Step 2 until $|\boldsymbol{\phi_s}|=K$ or $|\boldsymbol{\phi_r}|=\emptyset$
\end{itemize}

The complete procedure is formalized in Algorithm \ref{alg:proximity_sampling}.

\paragraph{Simulation and Analysis}

To validate the effectiveness of proximity sampling, we conduct a simulation experiment before deployment to investigate whether the data samples obtained through proximity sampling could effectively restore the rating scores of the models. Specifically, given a fixed total sampling budget $\mathcal{C}$, proximity sampling prioritizes allocating limited battle resources to model pairs with closely matched rating scores $\mathbf{u}$, whereas uniform sampling distributes comparisons randomly. 
\textbf{Simulation Procedure}:
\begin{itemize}[leftmargin=2em]
    \item \textbf{Initialization:} The simulation starts with a  predefined set of model rating scores \( \mathbf{u} \) and a threshold \( h \).  
    \item \textbf{Proximity sampling strategy:} Simulate a specified number of matchups by exclusively selecting model pairs within the threshold \( |u_i - u_j| < h \). 
    \item \textbf{Uniform sampling strategy:} Simulate matchups by randomly selecting model pairs. If we set \( h \geq u_{\text{max}} - u_{\text{min}} \), proximity sampling degenerates into uniform sampling.  
\end{itemize}
For both strategies, we generate pairwise corresponding comparison datasets. The outcome of each pairwise comparison is sampled probabilistically according to the win-rate formula in Eq. 3. These datasets are captured in two key matrices in order to visualize and interpret  the results:
\begin{itemize}[leftmargin=2em]
    \item \textbf{Payoff Matrix:} This matrix quantifies the relative performance between models. Each entry $(i, j)$ represents the win rate of model $i$ over model $j$.
    \item \textbf{Count Matrix:} This matrix records the distribution of pairwise comparisons. Each entry $(i, j)$ indicates the total number of comparisons conducted between model $i$ and model $j$. This matrix is crucial for assessing the statistical confidence of the win rates; a higher comparison count implies a more reliable estimate.
\end{itemize}
Following the methodology of Chatbot Arena, we estimate the rating scores from both datasets using the Bradley-Terry model with maximum likelihood estimation (MLE). The estimated scores are then compared against the ground-truth ratings $\mathbf{u}$. 

\begin{figure}[h]
    \centering
    \begin{subfigure}[b]{0.4\textwidth}
        \includegraphics[width=\textwidth]{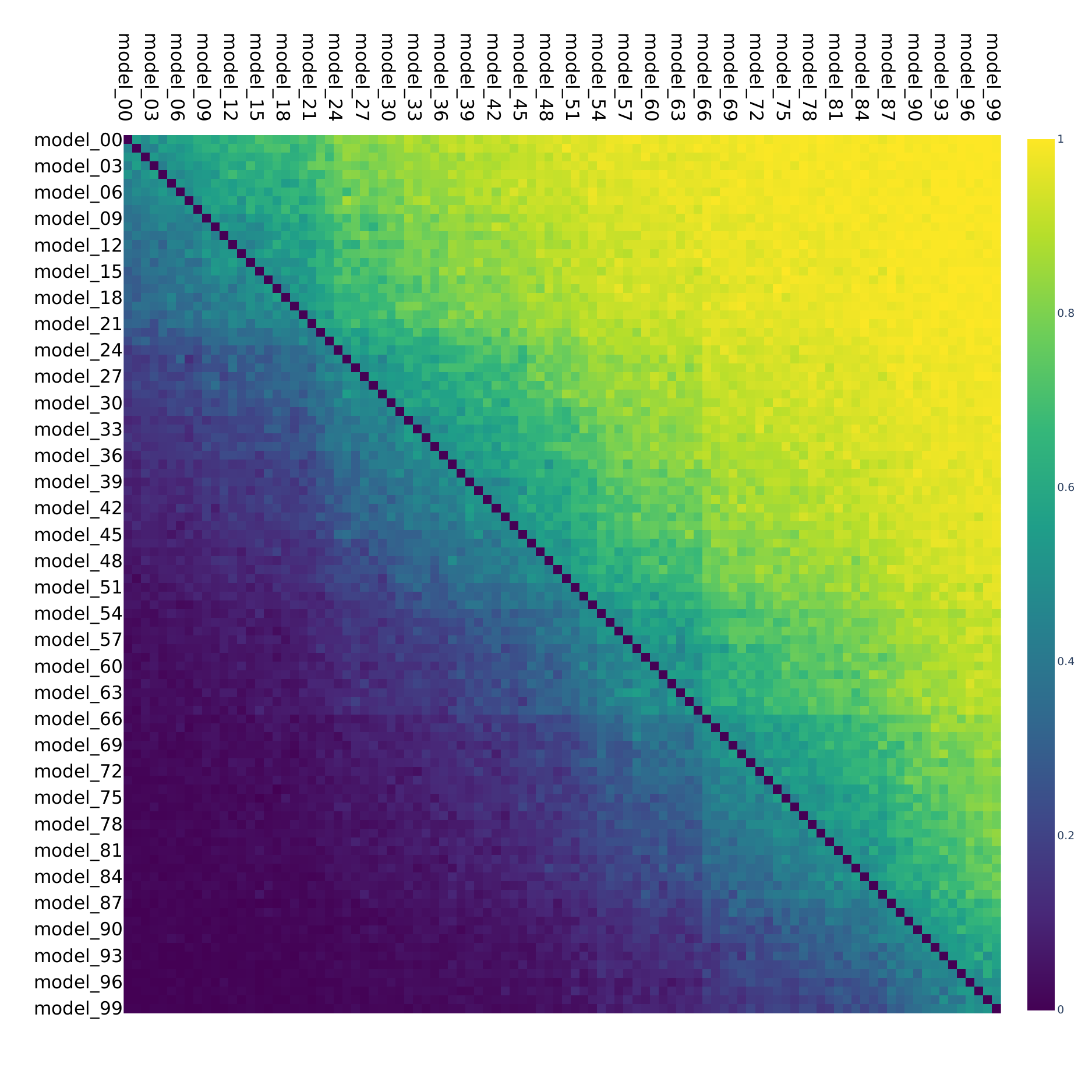}
        \subcaption*{(a) Uniform sampling: payoff matrix}
    \end{subfigure}
    \hspace{0.02\textwidth}
    \begin{subfigure}[b]{0.4\textwidth}
        \includegraphics[width=\textwidth]{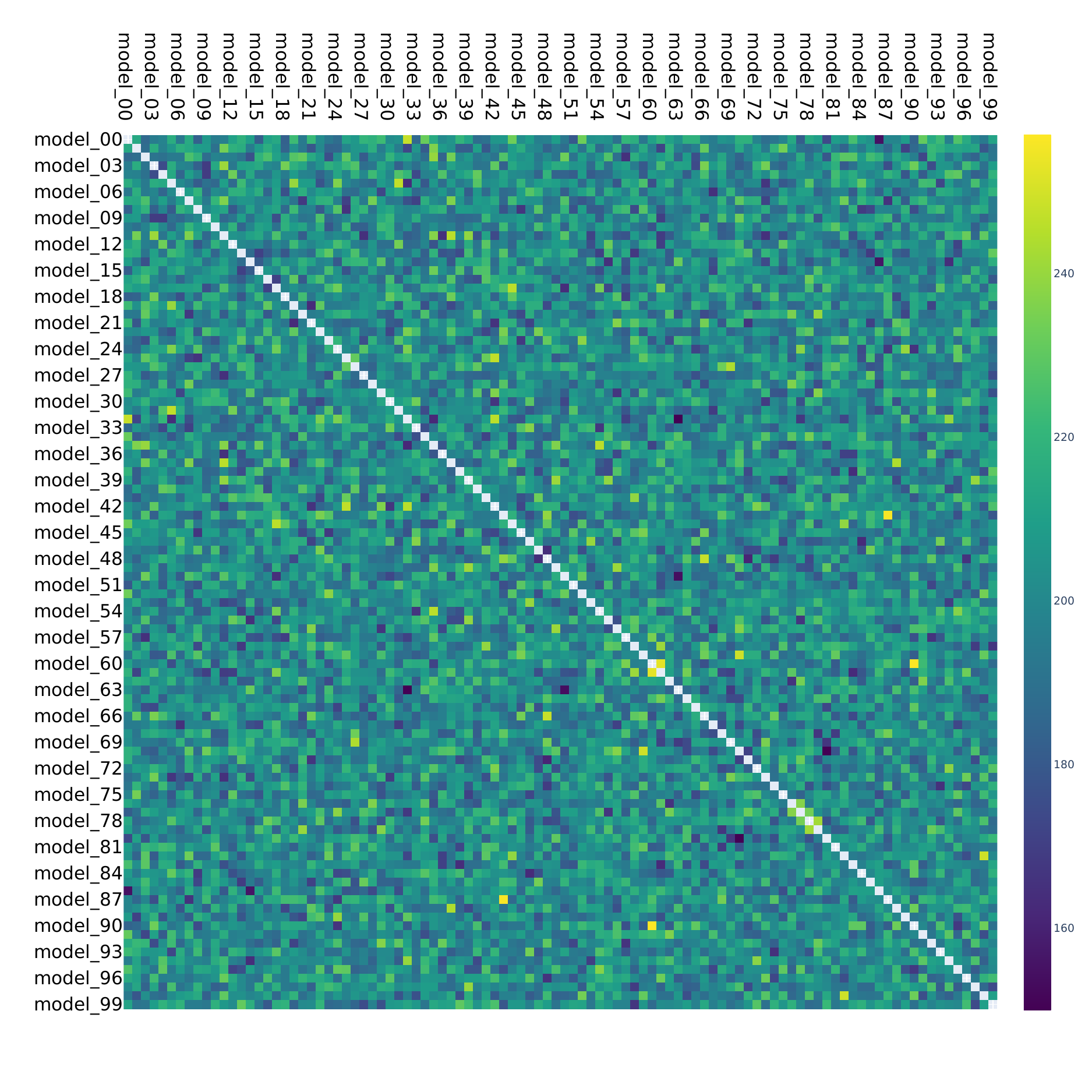}
        \subcaption*{(b) Uniform sampling: count matrix}
    \end{subfigure}
    \begin{subfigure}[b]{0.4\textwidth}
        \includegraphics[width=\textwidth]{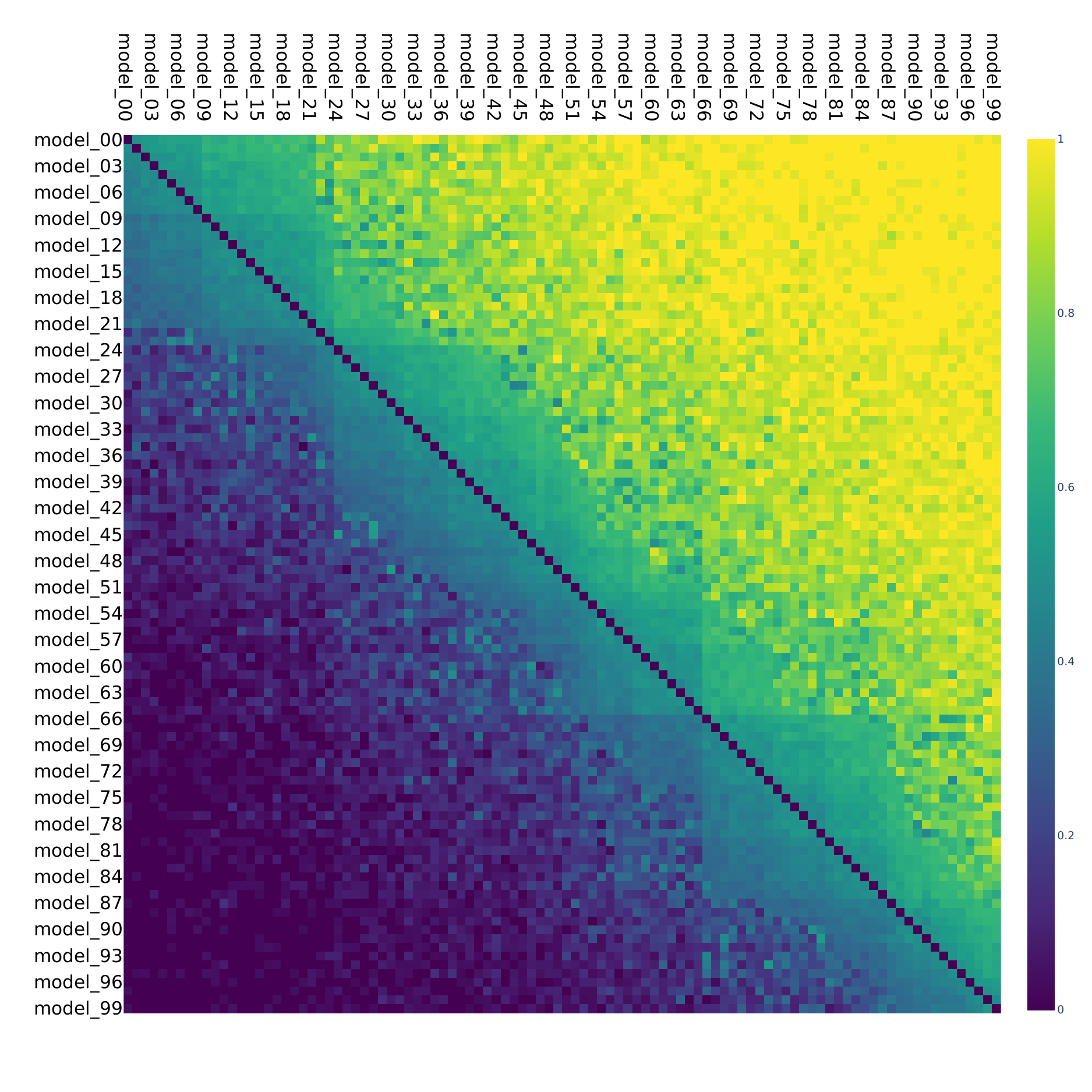}
        \subcaption*{(c) Proximity Sampling: payoff matrix}
    \end{subfigure}
    \hspace{0.02\textwidth}
    \begin{subfigure}[b]{0.4\textwidth}
        \includegraphics[width=\textwidth]{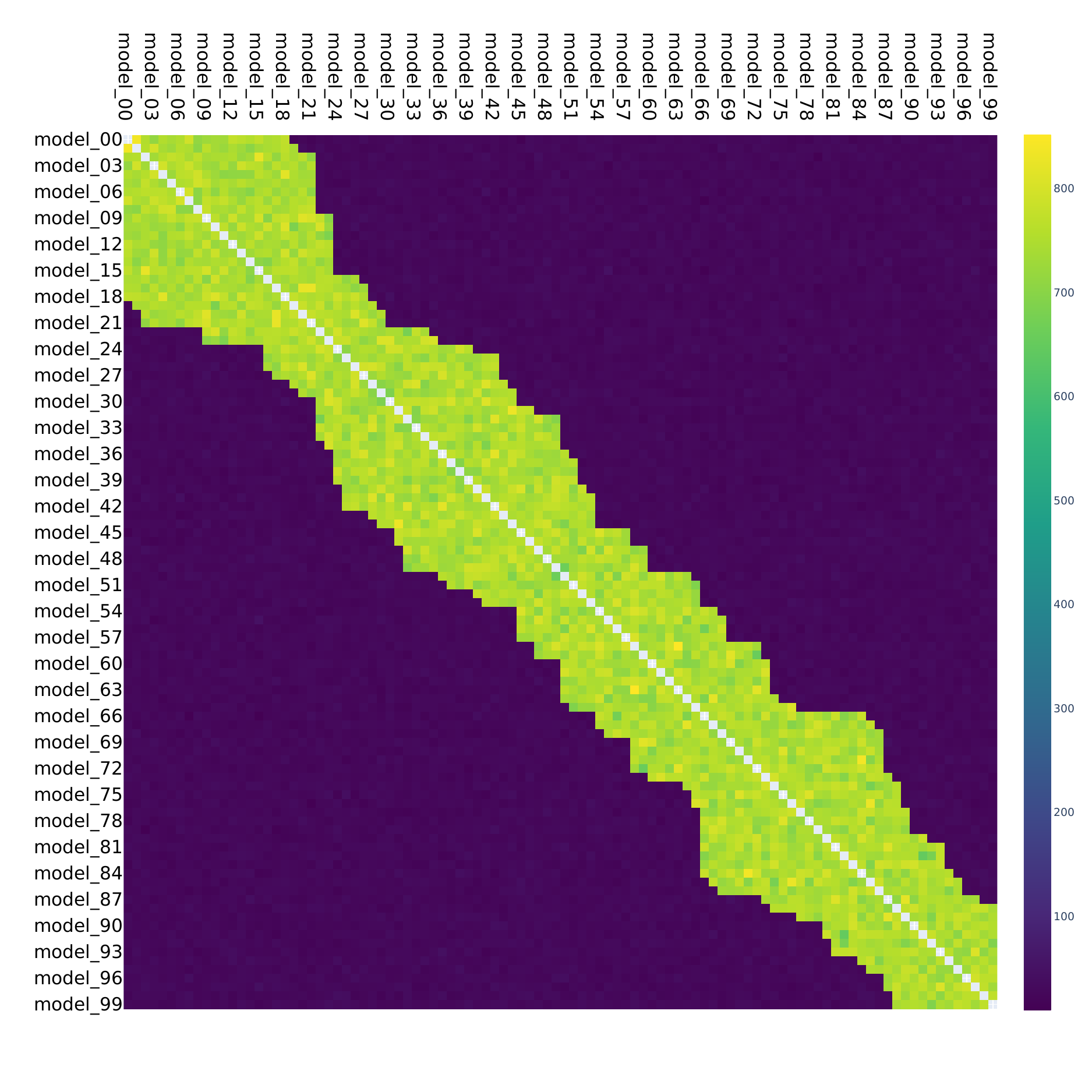}
        \subcaption*{(d) Proximity Sampling: count matrix}
    \end{subfigure}
    \caption{The payoff matrix and count matrix for proximity sampling (threshold = 120) and uniform sampling (threshold = 1000) from the simulation process. We add some random sampling to simulate non-proximal comparisons in real-world scenarios. The majority of comparisons are concentrated within each model's proximity region, so the count matrix appears as a band matrix. }
    \label{fig:simu_payoff}
\end{figure}

The payoff matrices and the count matrices obtained from the simulations of proximity sampling and uniform sampling are visualized in Figure \ref{fig:simu_payoff}. 
In these matrices, models are typically ordered from top to bottom (or left to right) according to their rating scores, from highest to lowest. The payoff matrix generated by proximity sampling approximates a band matrix: entries are densely concentrated near the diagonal but sparse overall. Uniform sampling, in contrast, produces a fully sparse matrix with no structural pattern. The underlying mechanism of proximity sampling relies on a progressive comparison chain: even if two models (e.g., A and Z) have a large Elo score gap and are never directly compared, \textbf{the overall comparability graph remains connected} as long as there exists a sequence of intermediate models.


    

\begin{figure}[h]
    \centering
    \begin{subfigure}[b]{0.45\textwidth}
        \includegraphics[width=\textwidth]{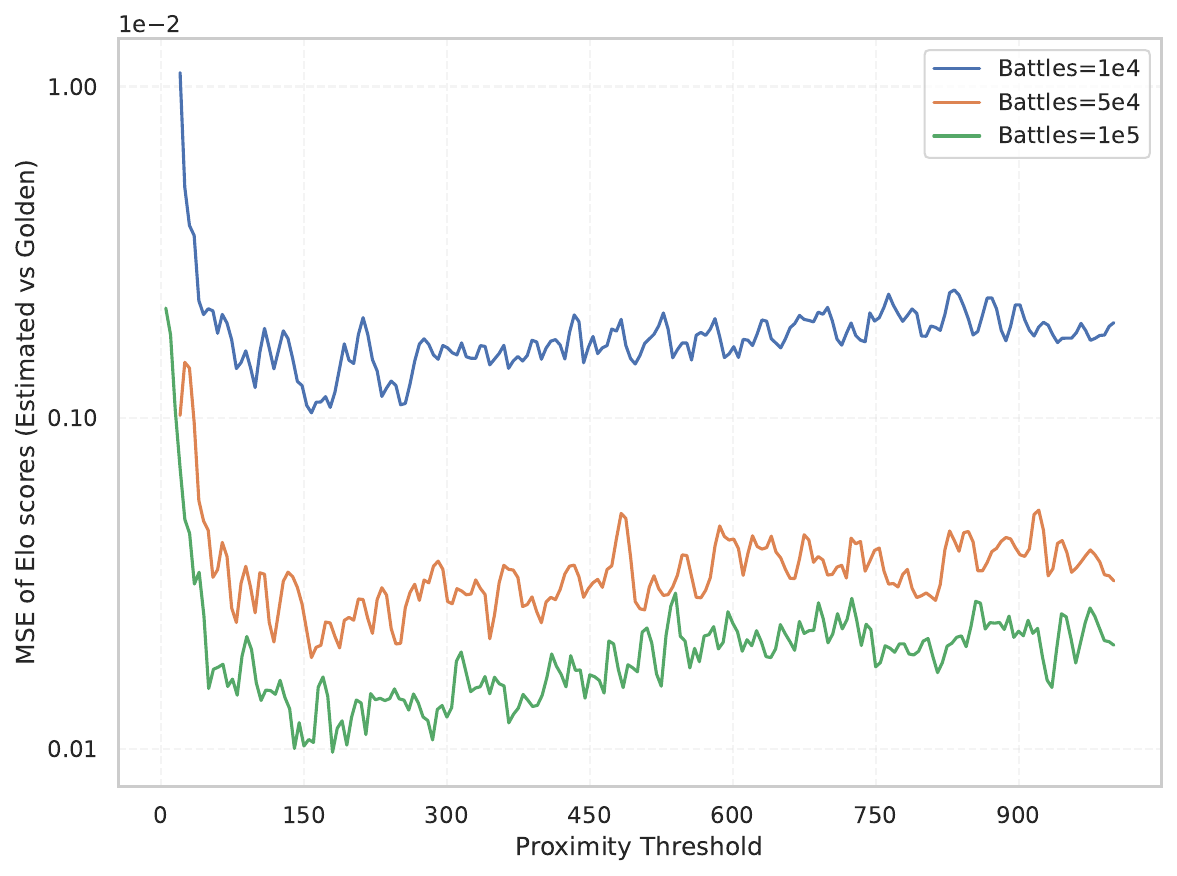}
        \subcaption*{(a) MSE between Estimated and Golden Elo Scores.}
        \label{fig:elo1}
    \end{subfigure}
    \hspace{0.01\textwidth}
    \begin{subfigure}[b]{0.45\textwidth}
        \includegraphics[width=\textwidth]{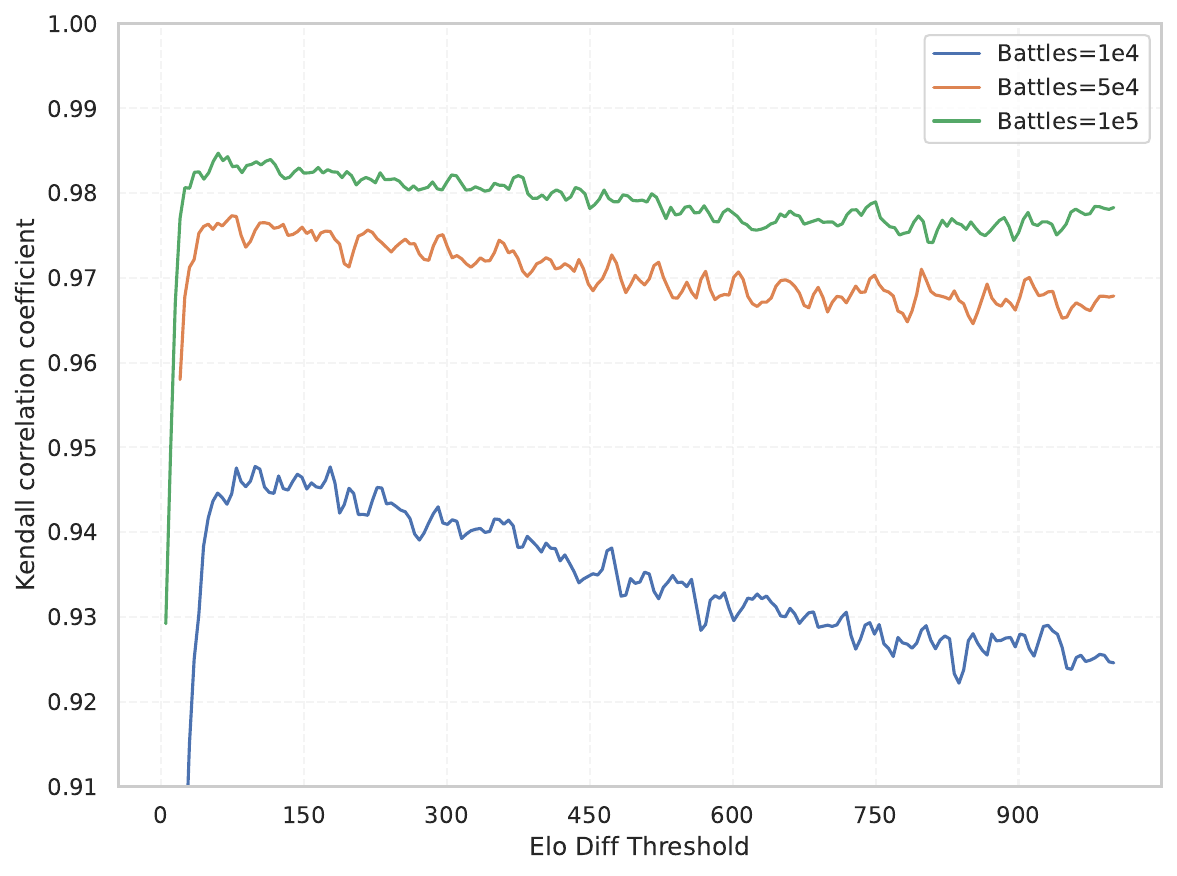} 
        \subcaption*{(b) Kendall correlation coefficient.}
        \label{fig:elo2}
    \end{subfigure}     
    \caption{Comparison of rating reconstruction performance under different total battle numbers and proximity thresholds. The x-axis represents the proximity threshold, where a larger value indicates a more relaxed sampling constraint. Each curve represents a different total battle number setting.}
    \label{fig:simu_res}
\end{figure}

We configure the simulation experiment with 100 simulated models and total sampling budgets of $\mathcal{C}=1e4, 5e4, 1e5$ to examine both resource-limited and sufficient sampling scenarios. Each model is assigned a golden arena rating score \( \textbf{u}^g_i \in (400, 1400) \). The threshold for proximity sampling is varied within (0, 1000), when it equals 1000, proximity sampling degenerates into uniform sampling. To better emulate real-world conditions where data may not strictly adhere to predefined constraints, we introduce noise into the simulated comparison data, allowing some samples to be drawn from outside the threshold range. 

From the simulated battle outcomes, we predict model ratings and arena scores (Elo scores) \( \mathbf{\hat{u}} \) via the BT Model and the MLE Algorithm. We evaluate the rating reconstruction performance using Root Mean Squared Error (RMSE) and Kendall's Tau (Kendall Correlation Coefficient). The simulation results are presented in Figure \ref{fig:simu_res}. The results demonstrate that when the proximity threshold is set around 150, RMSE reaches its minimum while Kendall's Tau peaks. This indicates that proximity sampling with this threshold produces data distributions that most accurately recover the true model capabilities.

We also conduct simulations on Proximity Sampling using the Chatbot Arena dataset in Appendix \ref{sec:simu_chatbot_arena}.

\subsection{Data Collection and Filtering}
We employ placement matches and proximity sampling for model battles in the Inclusion Arena Platform, enabling large-scale collection of user preference data on LLMs from real-world applications. To improve LLM ranking, further data filtering and analysis are necessary.

\paragraph{Data Collection} 
In apps integrated with the Inclusion Arena Platform, model battles are intermittently inserted during user-LLM interactions. 
In each battle round, given a conversation context, we sample $K$ \textbf{anonymous models} to generate responses. The user is then asked to select a winner. To minimize user burden, $K$ is currently set to 2 or 3, and each battle round yields $K-1$ pairwise model comparisons. In the early stages of the Inclusion Arena leaderboard, with no existing model rankings, placement matches and proximity sampling algorithms were not directly applicable at this stage. Therefore, for the first few models (approximately 10), we only used random sampling and collected initial data to rank models. Once initial rankings stabilized, we activated placement matches and proximity sampling. In this paper, we extract preference data from the Inclusion Arena Platform over a specific period, encompassing battles among 42 distinct models.

\paragraph{Data Filtering} 
To ensure the reliability of model evaluation and ranking, we filter pairwise comparisons by removing invalid data, including failed or rejected responses. Failed responses result from network errors or user misoperations, manifesting as garbled text, empty strings, or incomplete outputs. Rejected responses occur when LLMs decline to provide meaningful answers (e.g., "Sorry, I can’t help you."). Furthermore, to ensure compliance, we only retain data explicitly authorized by users and fully anonymize all collected data.


\begin{figure}[h]
    \centering
    \begin{subfigure}[]{0.99\textwidth} 
        \vspace{0.01\textwidth}
        \includegraphics[width=\textwidth]{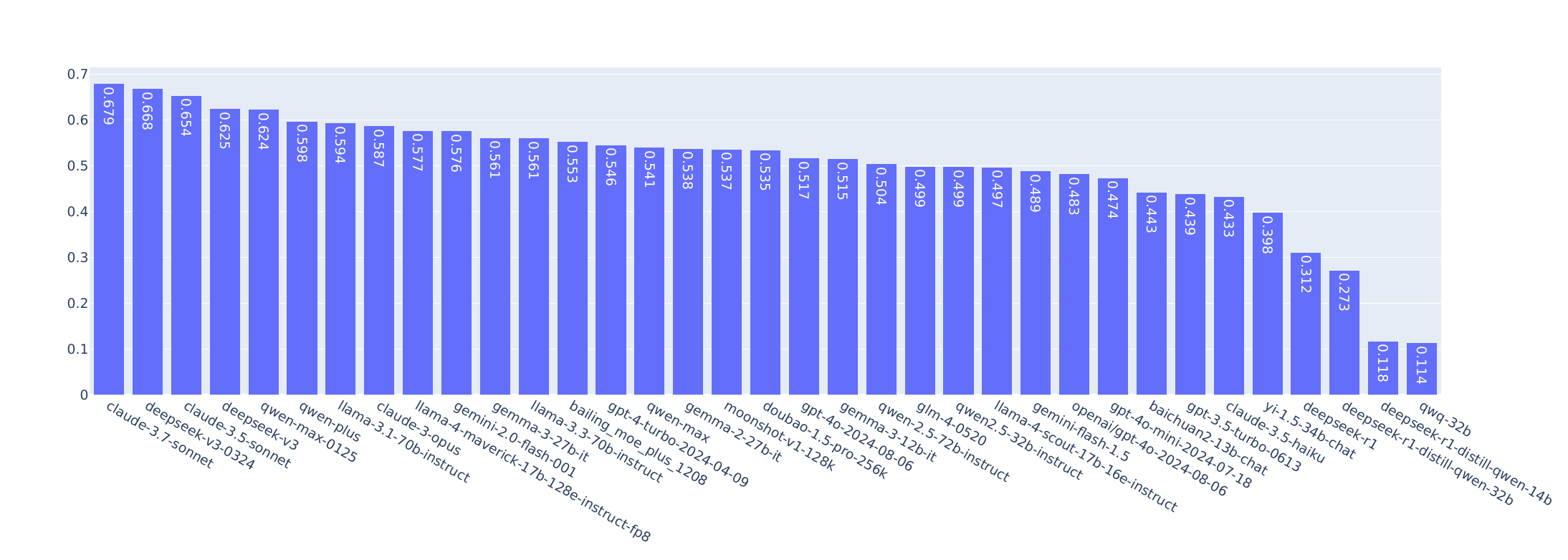}
        \subcaption*{(a) Average win rate per model}
    \end{subfigure}
    \vspace{0.01\textwidth}
    \begin{subfigure}[]{0.95\textwidth} 
        \includegraphics[width=\textwidth]{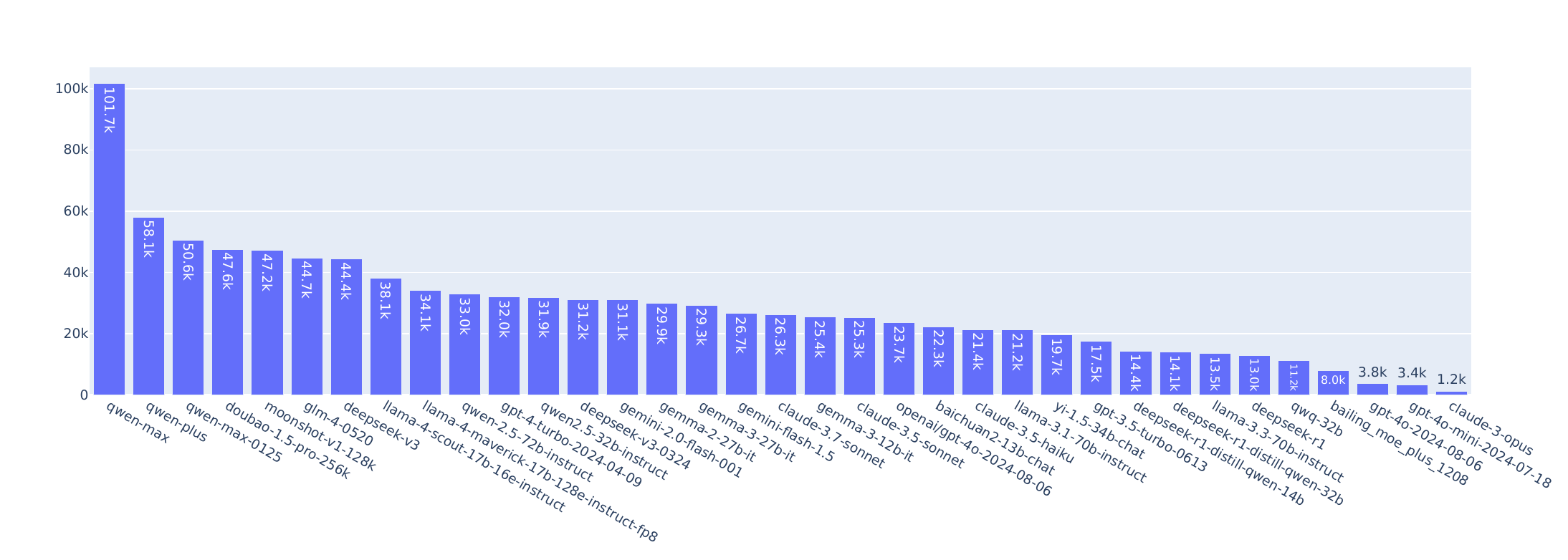}
        \subcaption*{(b) Total battle count per model}
    \end{subfigure}
    \caption{Distribution of per-model statistics based on collected pairwise comparisons.}
    \label{fig:per_model_stats}
    \end{figure}

\begin{table}[h]
\centering
\caption{The data statistics of pairwise comparisons collected from two integrated apps.}
\scalebox{0.85}{
\begin{tabular}{@{}lcccccc@{}}
\toprule
 & \# Pairs & \# Models & \# Users & Avg. Turn & Avg. Context length & Avg. Response length\\ \midrule
App1  & 493,094  &  49 & 45,170  &  9.31 & 2145.48 &  127.47 \\
App2  &   7,909  &  23 &  1,440  &  2.00 & 1407.33 &   81.08 \\ \midrule
Total & 501,003  &  49 & 46,611  &  9.19 & 2133.68 &  126.84 \\
\bottomrule
\end{tabular}}
\vspace{10pt}
\label{tab:statistics}
\end{table}

\subsection{Model Ranking}
Following Chatbot Arena~\citep{chatbot}, we employ the Bradley-Terry model with Maximum Likelihood Estimation to compute Inclusion Arena Scores, where $\alpha$ is set to $\ln(10)/400$ to maintain numerical consistency with the Elo rating scale. In this paper, we obtained a total of 501,003 pairwise comparisons for statistical evaluation and model ranking. The data is primarily sourced from two applications: \textit{Joyland} and \textit{T-Box}$^1$\footnote{$^1$The Joyland app delivers character-driven chats powered by LLMs, while the T-Box app serves as a practical AI agent platform that provides solutions and tools for real-world problems in daily life. Our data collection complies with data license agreements and explicit user consent.}.

\begin{figure}[ht]
    \centering
    \begin{subfigure}[b]{0.49\textwidth}
        \includegraphics[width=\textwidth]{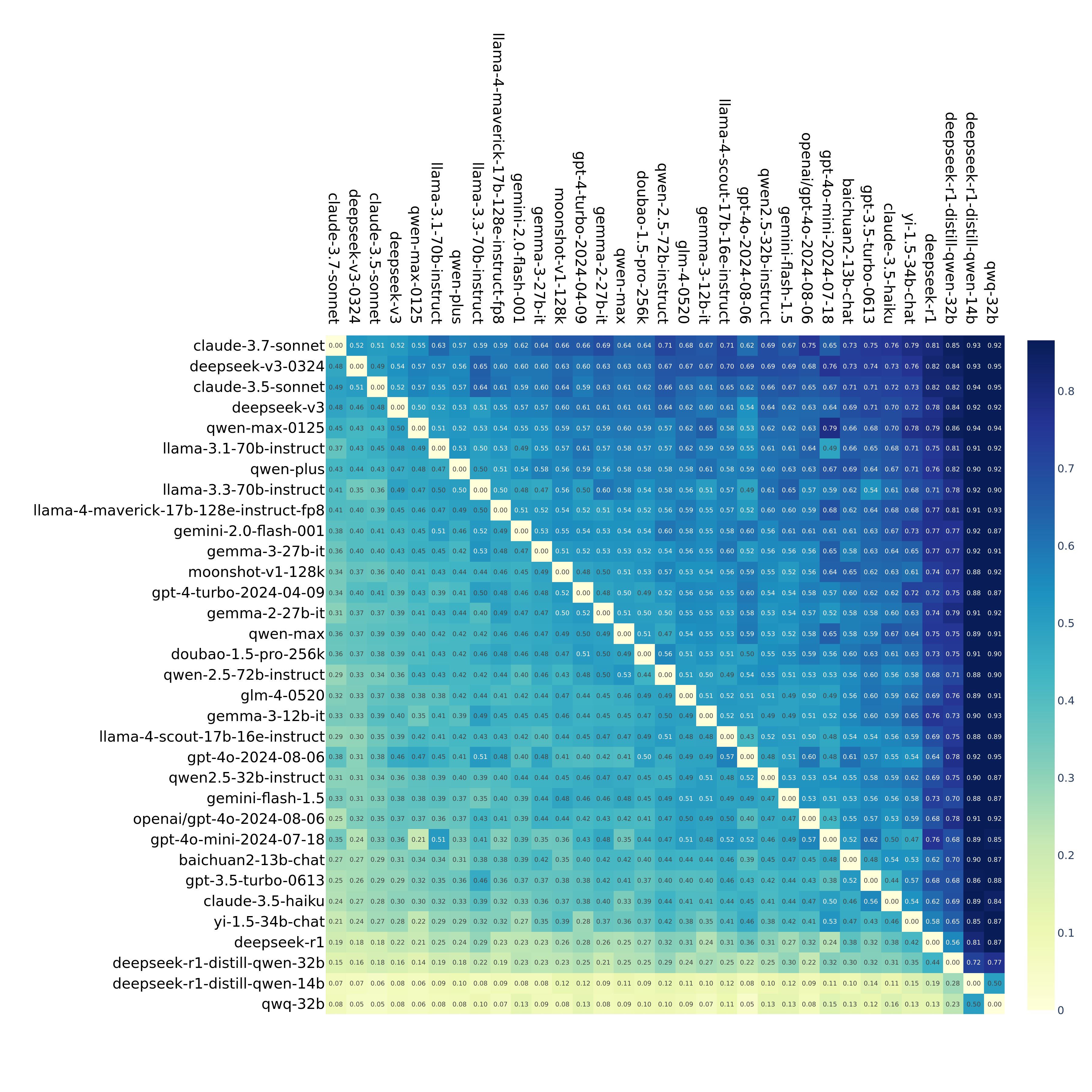}
        \subcaption*{(a) payoff matrix}
    \end{subfigure}
    \hspace{0.01\textwidth}
    \begin{subfigure}[b]{0.49\textwidth}
        \includegraphics[width=\textwidth]{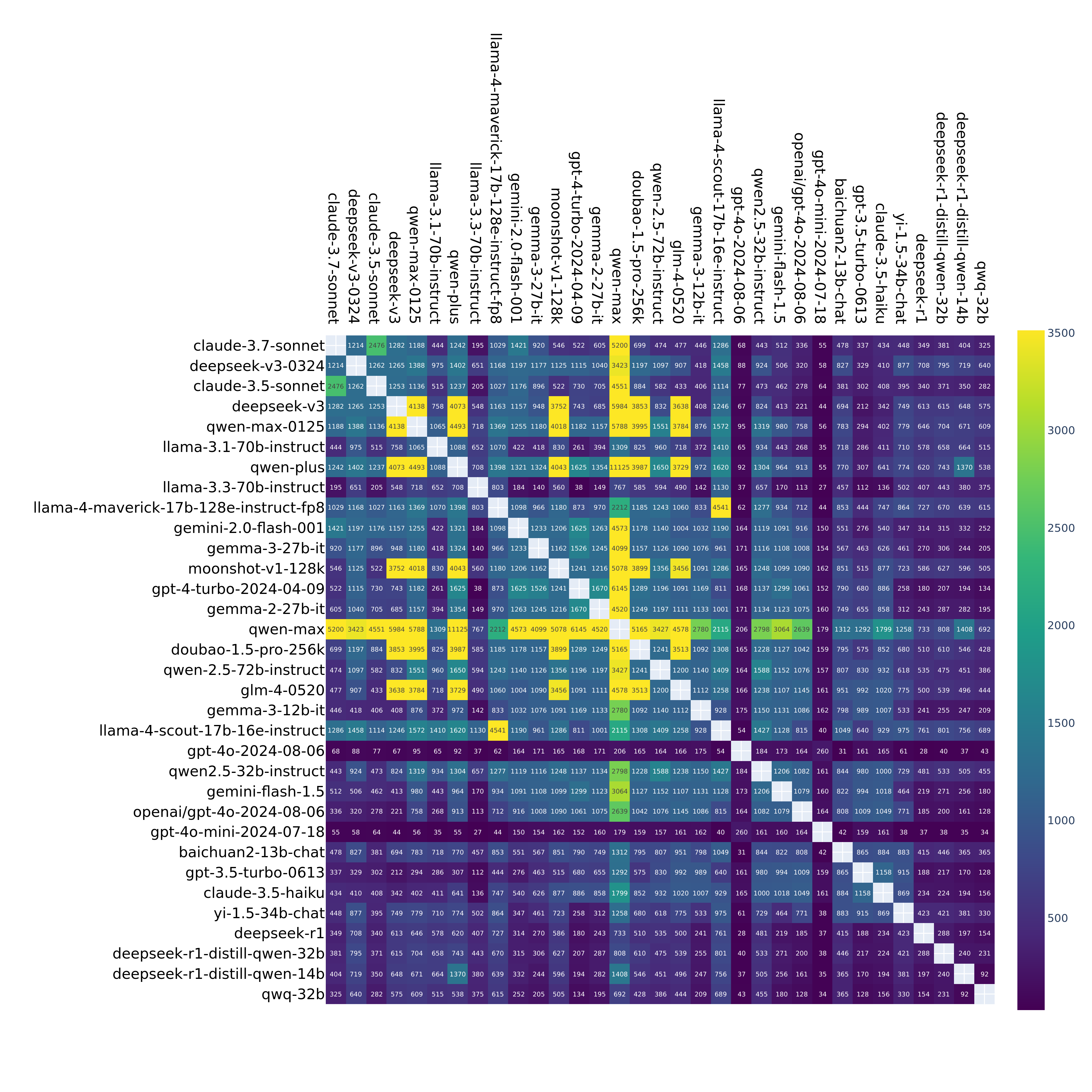}
        \subcaption*{(b) count matrix}
    \end{subfigure}
    \caption{Statistics of total collected pairwise comparisons from Inclusion Arena Platform.}
    \label{fig:iai_payoff}
\end{figure}

\begin{figure}[ht]
    \centering
    \begin{subfigure}[b]{0.49\textwidth}
        \includegraphics[width=\textwidth]{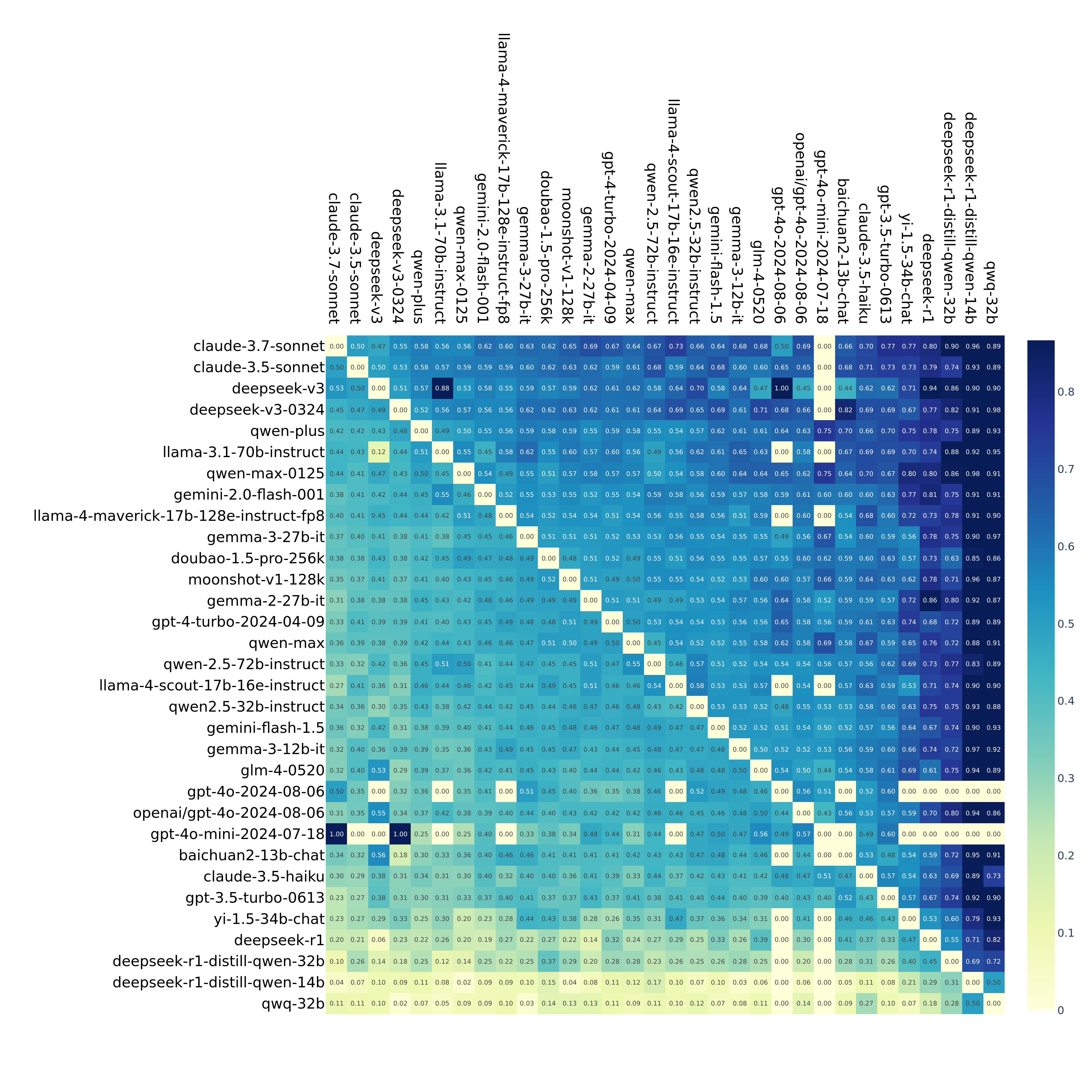}
        \subcaption*{(a) payoff matrix}
    \end{subfigure}
    \hspace{0.01\textwidth}
    \begin{subfigure}[b]{0.49\textwidth}
        \includegraphics[width=\textwidth]{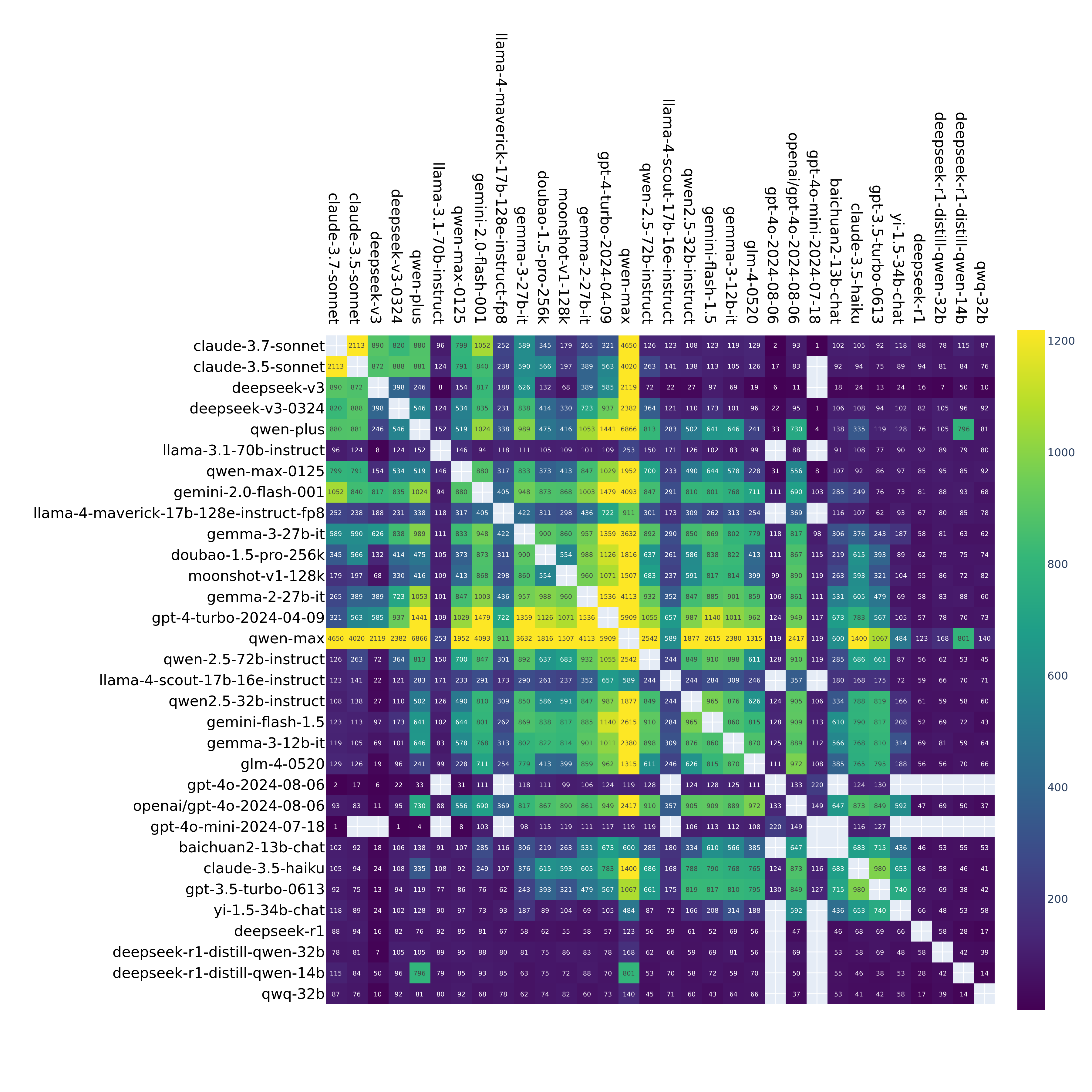}
        \subcaption*{(b) count matrix}
    \end{subfigure}
    \caption{Statistics of collected pairwise comparisons from Inclusion Arena Platform after applying Placement Matches and Proximity Sampling Algorithm.}
    \label{fig:iai_payoff_sample}
\end{figure}

During the establishment and operation of the Inclusion Arena platform, we adopted a phased approach to ensure system stability. In the initial phase (approximately two weeks), we limited the number of deployed models and employed random sampling. After achieving stabilized baseline model rankings, we progressively introduced additional models while implementing proximity sampling and placement matches algorithms. Additionally, to handle occasional API failures, we implement Qwen2.5-Max as a fallback model in the engineering pipeline when API outages occur. Table \ref{tab:statistics} summarizes the complete comparison dataset statistics. Different applications exhibit diverse usage patterns: tool-oriented scenarios typically involve fewer conversational turns, while daily conversations tend to have significantly more turns. Figure \ref{fig:iai_payoff} displays the payoff matrix and battle count matrix for a representative subset of evaluated models. Figure \ref{fig:per_model_stats} further illustrates each model's detailed win rate and corresponding battle count. Influenced by the random sampling during the platform's initial phase, the comparisons in the placement matches stage and the fallback mechanism, certain models exhibit a higher frequency of pairwise comparisons.  


Figure \ref{fig:iai_payoff_sample} presents the comparison data collected for selected models after the Inclusion Arena Platform implemented the Proximity Sampling algorithm following its initial phase. The battle count matrix demonstrates that when excluding sparse instances (primarily from placement matches and API failures), the empirical data distribution closely approximates a band matrix structure under Proximity Sampling. The number of models will progressively increase over time (see Figure \ref{fig:temporal_stats}), while unsuitable models will be systematically phased out based on App requirements and user preferences. As the model pool expands, the efficiency advantages of proximity sampling become increasingly pronounced. As detailed in our bootstrap analysis in Appendix \ref{sec:bootstrap_analysis}, the Elo ratings derived from proximity-sampled data exhibit substantially lower variance than those from uniformly sampled data, confirming the robustness of our approach.

\section{Discussions}

\subsection{Theoretical Analysis of Proximity Sampling}\label{Discussion:theoretical}

To accurately estimate the skill ratings $\theta = (u_1,...,u_n)^T$ for a set of models, we adopt the Bradley-Terry (BT) model, where the probability of model $i$ defeating model $j$ in $\iota$-th match is given by $p_\iota = \sigma(\alpha(u_{i} - u_{j}))$, where $\sigma(\cdot)$ is the logistic function, and $\alpha$ scales the skill difference.   Since each match outcome $y_\iota$ is a Bernoulli random variable, the model ratings are typically estimated using Maximum Likelihood Estimation (MLE) by maximizing the log-likelihood function:
\begin{equation}
\ell(\theta) = \sum_{\iota=1}^\mathcal{C} \left[ y_\iota \log p_\iota + (1 - y_\iota) \log \left(1 - p_\iota \right) \right].
\label{eq:loglikelihood}
\end{equation}

A critical question arises: given a fixed total number of matches, how can we design matches to yield the most precise and reliable model ratings? The precision of these MLE estimates is fundamentally quantified by the Fisher Information Matrix (FIM), denoted $I(\theta)_{ij} = -\mathbb{E}\left[ \frac{\partial^2 \ell(\theta)}{\partial u_i \partial u_j} \middle| \theta \right]$. 
Specifically, for a single match \( \iota \) between players \( i \) and \( j \), its contribution to the Fisher information matrix \( I_\iota \) is given by:  
\begin{equation}
[I_{\iota}]_{ii} = [I_{\iota}]_{jj} = \alpha^2 p_{ij}(1 - p_{ij}), \quad [I_{\iota}]_{ij} = -\alpha^2 p_{ij}(1 - p_{ij}),
\end{equation}
The aggregate Fisher information matrix \( I(\theta) \) over all matches can be expressed as:  
\begin{equation}\label{equ:i_theta_details}
I(\theta) = \sum_\iota I_\iota = \sum_{i<j} c_{ij} \alpha^2 f_{ij} A_{ij},
\end{equation}
where \( c_{ij}  \) is the number of comparisons between players \( i \) and \( j \),  \( f_{ij} = p_{ij}(1 - p_{ij}) \) is the variance term,  \( A_{ij} = (e_i - e_j)(e_i - e_j)^\top \) is a symmetric positive semidefinite matrix, with \( e_i \) being the standard basis vector.
By the \textbf{Cramér-Rao lower bound}, the variance of any unbiased estimator $\hat{\theta}$ is bounded by:
\begin{equation}
\text{Var}(\hat{\theta}) \succeq I(\theta)^{-1},  
\end{equation} 
where $\succeq$ denotes the Loewner order (i.e., $\text{Var}(\hat{\theta}) - I(\theta)^{-1}$ is positive semidefinite). 
Under large-sample asymptotics, the MLE is consistent and asymptotically normal, with the covariance matrix of $\hat{\theta}$ converging to the inverse of the FIM. For individual ratings $u_i$, this implies: 
\begin{equation}
\text{Var}(\hat{u}_i) \approx [I(\theta)^{-1}]_{ii}.
\end{equation}

Therefore, we use the trace of \( I(\theta)^{-1} \), denoted as \(\text{tr}[I(\theta)^{-1}]\), to measure the total variance of the estimator.
By defining the weights \( w_{ij} = c_{ij}f_{ij} \geq 0 \), the FIM can be expressed in terms of the weighted graph Laplacian matrix \(L(w)\): 
\begin{equation}
    L(w) := \sum_{i < j} w_{ij} (e_i - e_j)(e_i - e_j)^\top
\end{equation}
which we refer to as the weighted Laplacian matrix. This yields \( I(\theta) = \alpha^2 L(w) \) and consequently \( \text{tr}[I(\theta)^{-1}] = \alpha^{-2} \text{tr}[L(w)^{\dagger}] \). Since $L(w)$ is singular, $L(w)^{\dagger}$ denotes the Moore-Penrose pseudo-inverse of $L(w)$.

To further evaluate the behavior in the objective function $\text{tr}[I(\theta)^{-1}]$, we conduct numerical simulations by sweeping the proximity threshold $h$ over $[100,1000]$. We consider 100 models with ratings sampled from the ELO rating space $(0,1000]$, with the total battle count $\mathcal{C}$ fixed at $10^4, 10^5$, and $10^6$. The "ideal" case assumes uniformly distributed model capabilities with equal battle counts for all proximity model pairs, while "practical" corresponds to the simulation in Section \ref{sec:proximity_sampling}, where model capabilities are randomly pre-defined and proximity model battle counts are sampled via Algorithm \ref{alg:proximity_sampling} to approximate uniformity.

\begin{figure}[ht]
    \centering
    \begin{subfigure}{0.49\textwidth}
        \includegraphics[width=\textwidth]{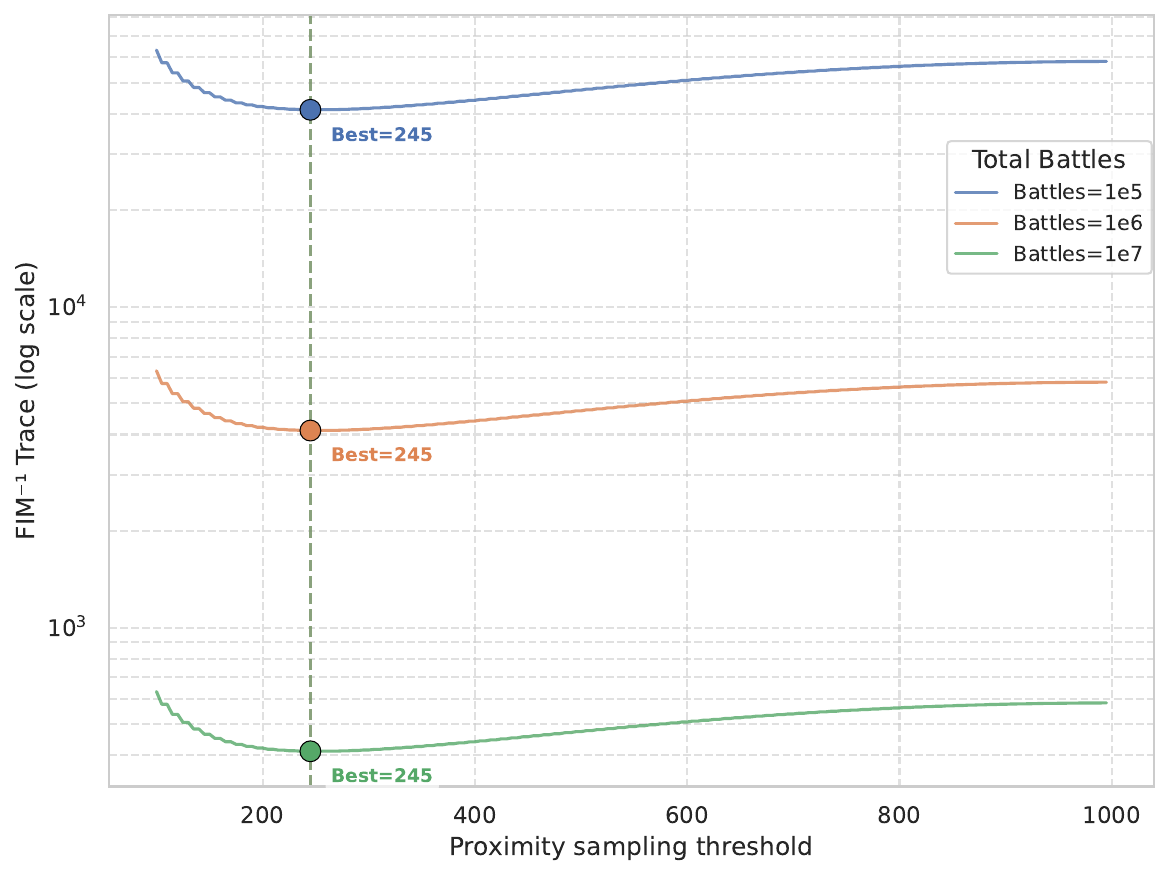}
        \subcaption*{(a) ideal}
        \label{fig:simu0_a}
    \end{subfigure}
    \begin{subfigure}{0.49\textwidth}
        \includegraphics[width=\textwidth]{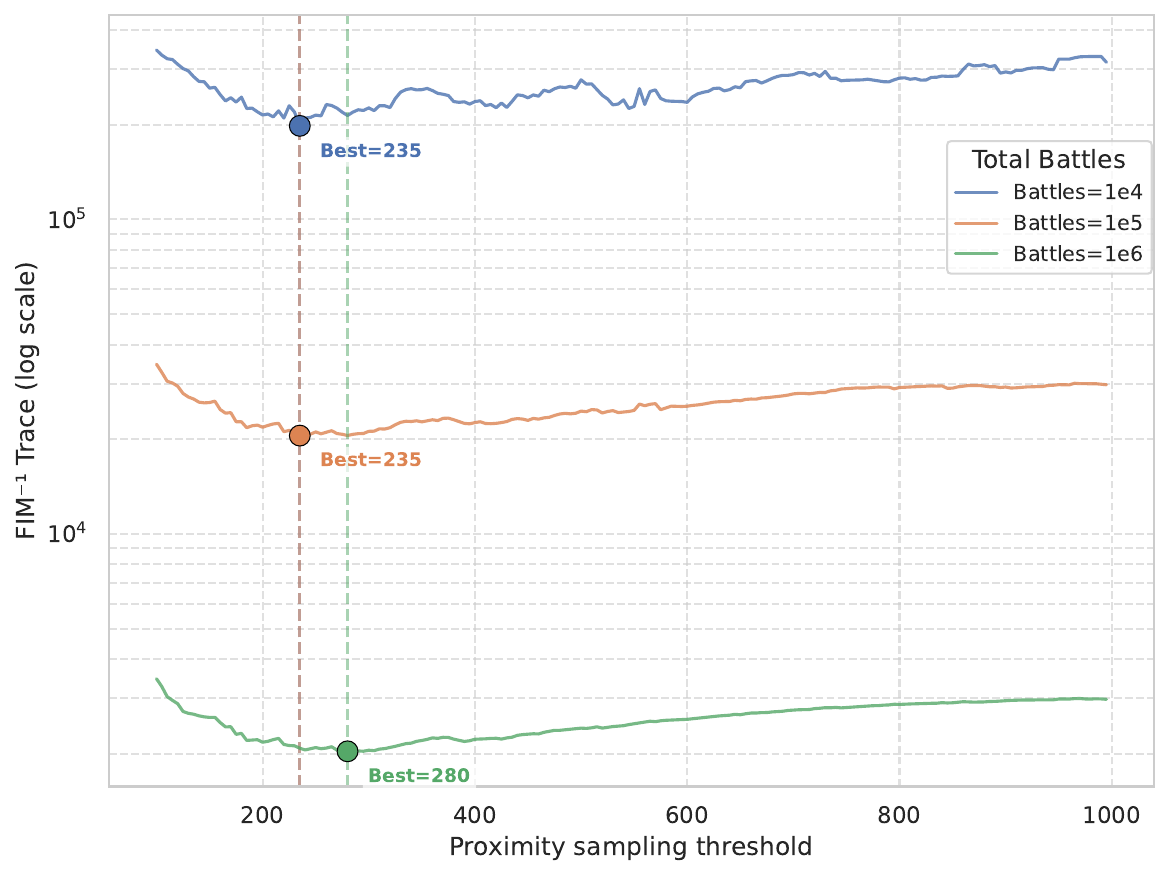}
        \subcaption*{(b) practical}
        \label{fig:simu0_b}
    \end{subfigure}
    \caption{The plot illustrates the relationship between this proximity sampling threshold ($h$) and the total variance of model ELO estimates, represented by the trace of the inverse Fisher Information Matrix (FIM).} 
    \label{fig:simu0}
\end{figure}

Our numerical simulations reveal three key findings. \textbf{First,} $\text{tr}[I(\theta)^{-1}]$ exhibits a \textbf{U-shaped trend} w.r.t. proximity threshold $h$ (Figure \ref{fig:simu0}), with optimal performance achieved at intermediate thresholds. We attribute this to a \textbf{trade-off}: small values of $h$ may result in an insufficiently connected comparison graph, while large values of $h$ introduce less informative comparisons (i.e., pairs with deterministic outcomes), thereby degrading estimation efficiency. Notably, proximity sampling consistently outperforms random sampling ($h{=}1000$) across all $\mathcal{C}$, which highlights a key insight: For a fixed model set and total comparison count, there exists a specific proximity threshold interval that outperforms random sampling. \textbf{Second}, proximity sampling shows advantages in data-limited scenarios. In Figure \ref{fig:simu0}(b), at \( \mathcal{C}=10^4 \), it reduces \(\text{tr}[I(\theta)^{-1}]\) by 37.19\% compared to random sampling (\( h=1000 \)), narrowing to 31.65\% at \( \mathcal{C}=10^6 \), highlighting its information efficiency priority. \textbf{Furthermore,} while the ideal case suggests an optimal $h^*$, the practical case shows limited variability. Nevertheless, the robustness of proximity sampling over uniform sampling is evident, offering a reliable strategy for empirical applications. We provide further discussion in Appendix \ref{sec:further_discussion}.

\subsection{Data transitivity}
Data within the Inclusion Arena platform originates from real-world applications, where user-generated prompts are more aligned with authentic user experience (UX). This distinguishes its data distribution from that of crowdsourced platforms. We utilize open-source data from Chatbot Arena as a representative of crowdsourced data characteristics and compare it with randomly sampled data collected from Inclusion Arena. Following the Disc Decomposition algorithm~\citep{bertrand2023limitationselorealworldgames} and Equation~\ref{eq:disc_decom}, we fit the $(u,v)$ parameters for both datasets and visualize the results.

\begin{figure}[ht]
    \centering
    \begin{subfigure}[b]{0.45\textwidth}
        \includegraphics[width=\textwidth]{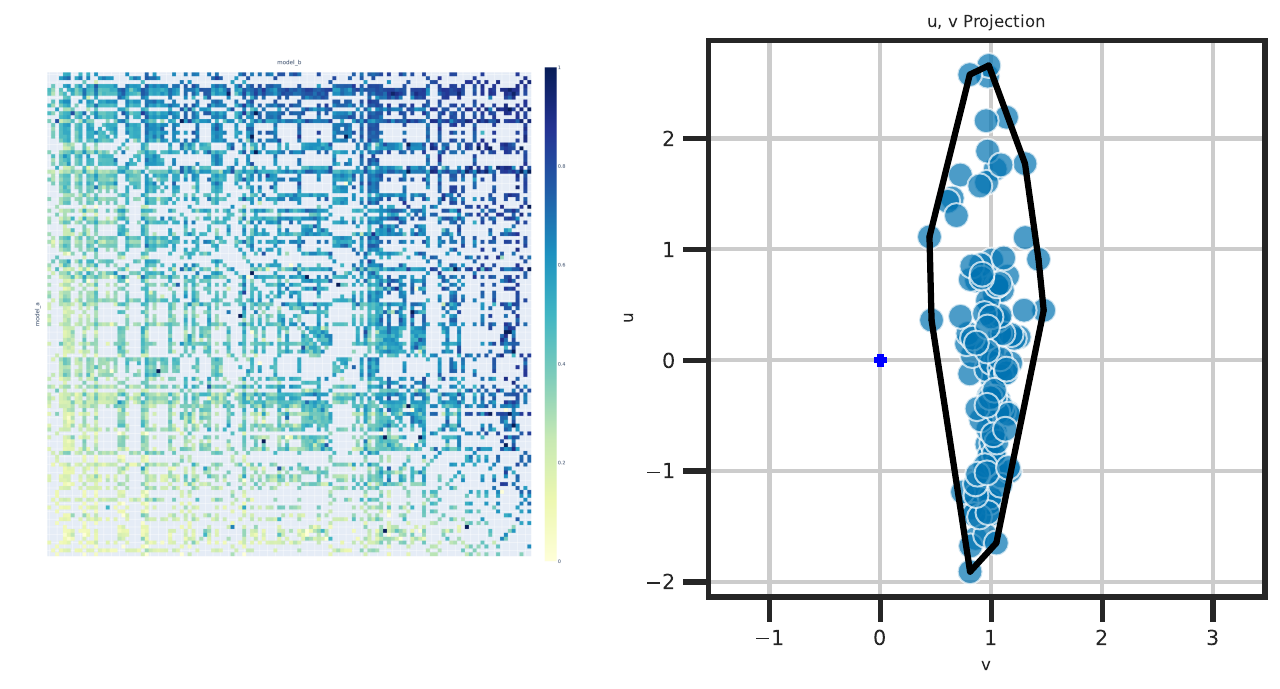}
        \subcaption*{(a) $(u,v)$ visualization for Chatbot Arena data}
    \end{subfigure}
    \hspace{0.02\textwidth}
    \begin{subfigure}[b]{0.45\textwidth}
        \includegraphics[width=\textwidth]{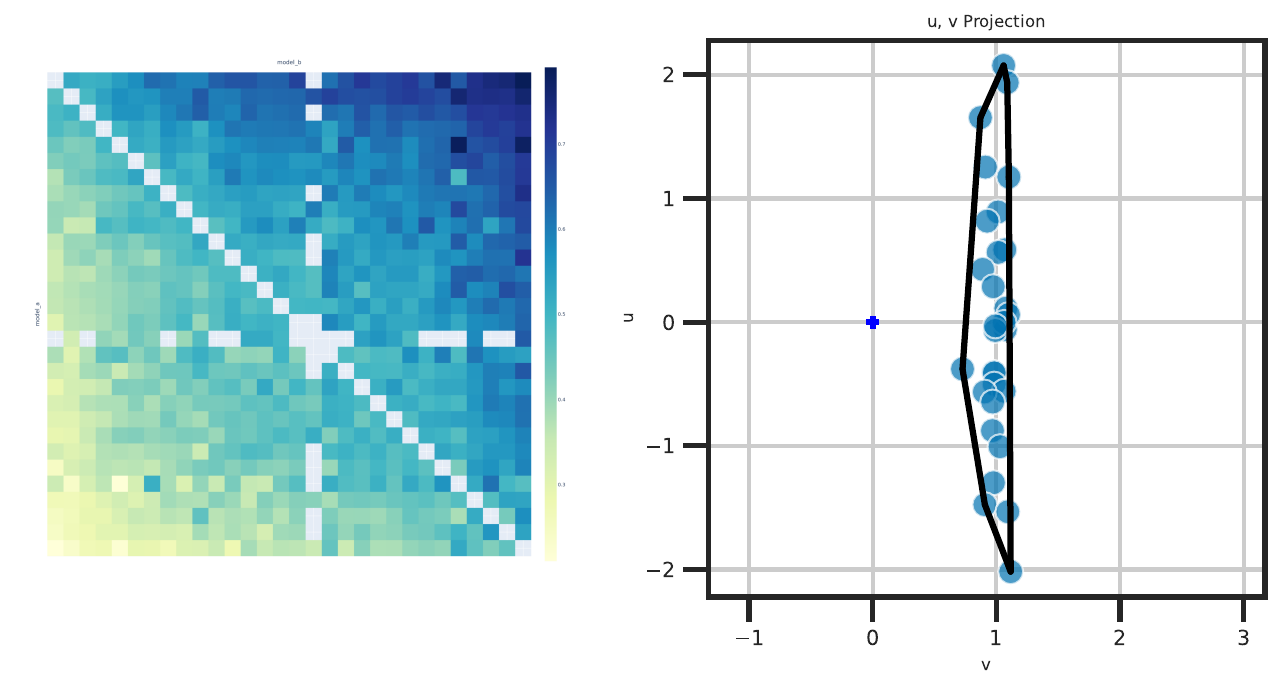}
        \subcaption*{(b) $(u,v)$ visualization for Inclusion Arena data}
    \end{subfigure}
    \caption{Visualization of the Disc Decomposition $(u,v)$ parameters. To ensure statistical reliability and avoid potential bias from insufficient samples, we only retain model pairs with more than 2,000 pairwise comparisons. }
    \label{fig:iai_disc_decompositon}
\end{figure}

The $(u, v)$ visualizations for both Chatbot Arena and Inclusion Arena data are presented in Figure \ref{fig:iai_disc_decompositon}. In particular, the data from Inclusion Arena exhibit $v_i$ values close to 1 and show a smaller variance compared to the Chatbot Arena data. According to the analysis in~\citep{bertrand2023limitationselorealworldgames}, this configuration suggests that the game structure is closer to an Elo-like game, where transitivity is high and $u_i$ serves as the dominant axis. Conversely, the $(u_i, v_i)$ pairs derived from Chatbot Arena data are more spread out, indicating characteristics closer to a Disc game with potential cyclic relationships. This suggests that for battles on the Inclusion Arena platform, which are more reflective of user experience, the Bradley-Terry (BT) and Elo models are likely to offer greater reliability for fitting model rankings due to the higher transitivity observed in the data.

\subsection{Building a Reliable Evaluation Platform}

The Inclusion Arena platform mechanisms and the applied placement and proximity sampling methodology offer significant advantages in terms of rating stability and system security. 

\paragraph{Stability} By prioritizing comparisons between models with uncertain outcomes, the rating system can more effectively capture fine-grained performance differences. This strategy promotes faster convergence to true skill ratings and yields more stable win-rate estimates. Additionally, the \textbf{connectivity of the comparison graph} plays a fundamental role in ensuring ranking reliability and stability. As demonstrated by \citet{singh2025leaderboardillusion}, when evaluating large numbers of models, limited comparison trials often result in a sparse count matrix (or a disconnected comparison graph), leading to distorted rankings that significantly diverge from ground truth performance. Theoretically, a disconnected comparison graph leads to a singular Fisher Information Matrix (FIM), making the relative rankings between disconnected sub-groups entirely unidentifiable. Our proximity sampling strategy addresses these challenges by explicitly optimizing the comparison graph structure. This approach not only guarantees graph connectivity but also reduces the total estimation variance (denoted as $\text{tr}[I(\theta)^{-1}]$), leading to more stable rankings while improving resource efficiency.

\paragraph{Security} Compared to other platforms, Inclusion Arena also enhances security against human manipulation. As noted by \citet{singh2025leaderboardillusion} and \citet{min2025improvingmodelrankingchatbot}, some evaluators on open platforms may intentionally submit erroneous results to distort rankings. They can even identify a model’s "identity" and strategically submit hundreds of votes to drastically alter its ranking. In contrast, manipulating scores on the Inclusion Arena Platform is significantly more difficult due to the following safeguards:
\begin{itemize}[leftmargin=2em]
    \item Our apps are designed for real users and typically require registration, establishing \textbf{a natural barrier against score manipulation.}
    \item Model battles are initiated randomly by the system rather than user choice, increasing the time cost for malicious actors—particularly in multi-turn conversations where battles are occasionally triggered.
    \item  By restricting comparisons to models with similar ability levels, attackers should generate a substantially larger number of distorted battle results to noticeably inflate a model’s ranking.
\end{itemize}


\begin{figure}[htbp]
    \centering
    \label{fig:category}
    \includegraphics[width=0.9\linewidth]{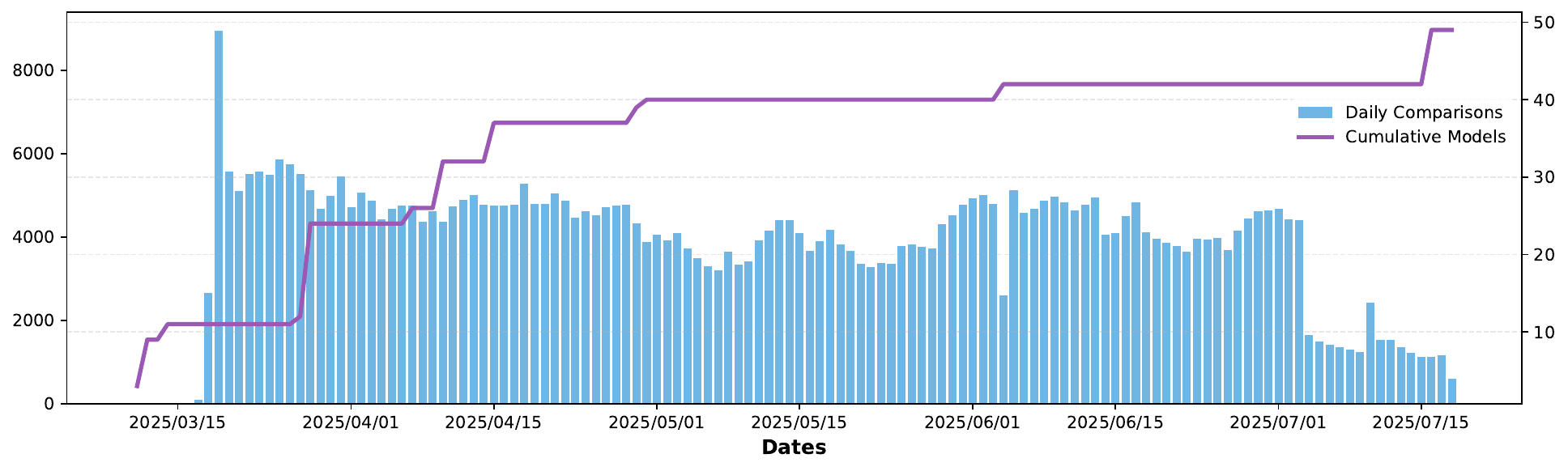}
    \caption{Temporal statistics of collected pairwise comparisons from Inclusion Arena Platform.}
    \label{fig:temporal_stats}
\end{figure}

\paragraph{Scalability} Figure \ref{fig:temporal_stats} illustrates the platform's efficient data acquisition capabilities and robust growth trend. Over a three-month observation period, the platform efficiently accumulated over 40,000 valid model comparison data points. During this time, the daily traffic volume remained at a consistently high and stable level, while the number of participating models also demonstrated steady growth.

This growth pattern is fundamental to the leaderboard's reliability. The substantial data provides a solid statistical foundation for high-confidence rankings. Meanwhile, the consistent data velocity ensures the leaderboard's timeliness, allowing it to dynamically reflect the most current model capabilities amidst evolving user preferences and model updates.

\subsection{Sub Ranking in Apps}\label{sec:sub_rank}
Currently, we focus on collecting data and ranking models on two apps, each catering to different user scenarios. Moving forward, the Inclusion Arena Platform will gradually support more apps. Across different apps, user demographics and preferences may vary significantly due to differences in app design philosophies and intended user experiences. As a result, not all models may be suitable for every app, and models deemed incompatible with a particular app may be removed. 

Given that the associated models for each app can differ substantially, and that each app emphasizes different fine-grained model capabilities. For instance, if Model A outperforms Model B in one app, this conclusion may not necessarily hold in another. To better serve users, we plan to introduce app-specific sub-leaderboards in the future. This finer-grained ranking approach will provide more tailored insights into model capabilities within each app’s unique context.

\section{Conclusion}
In this paper, we introduce \textbf{Inclusion Arena}, \textbf{a novel live leaderboard platform designed to bridge the gap between large foundation models (LLMs and MLLMs) and their practical utility in real-world AI-powered applications}. Our contributions are multi-faceted; we propose a systematic methodology for integrating model evaluation directly into user interactions within real-world applications, enabling the collection of highly relevant human preference data. To address critical challenges inherent in dynamic ranking,  we develop \textbf{Placement Matches} for efficient cold-start estimation of new models and \textbf{Proximity Sampling} to optimize information gain by prioritizing comparisons between models of similar capabilities. Empirical simulations and analyses of real-world data collected from Inclusion Arena demonstrate that our approach yields more stable and accurate rankings, while also enhancing security against data manipulation compared to general crowdsourcing platforms. By fostering an open alliance between foundation models and real-world applications, Inclusion Arena aims to accelerate the development of LLMs and MLLMs that are truly optimized for practical deployment and user experience. Future work will focus on expanding the integration of more diverse applications and exploring app-specific sub-leaderboards to provide finer-grained insights into model performance across varied user contexts. We hope Inclusion Arena could propel the AI community towards more robust and user-centric applications.

\section*{Limitations}
The Inclusion Arena platform is currently in its early stages, supporting only a limited number of apps. As the platform matures, we plan to gradually expand the range of integrated apps to cover diverse real-world scenarios. At present, model battles are not yet supported in multimodal settings due to operational and deployment costs. We will extend our evaluation framework to support multimodal models after the current system demonstrates stable performance. Furthermore, the existing model ranking is general-purpose and does not differentiate between application domains. We intend to introduce domain-specific sub-leaderboards (e.g., for education, entertainment) as well as app-specific sub-leaderboards to provide more nuanced evaluations. 
\bibliographystyle{antgroup}
\bibliography{ref/reference}

\clearpage
\appendix
\section{Further Discussion of Proximity Sampling }\label{sec:further_discussion}

In \textbf{proximity sampling} algorithm, we determine the proximity threshold $h$ and uniformly distribute the total number of battles $\mathcal{C}$ among the set of proximity model pairs. Let the \textbf{proximity model pair set} be defined as $S(h) = \{(i,j) \mid 1 \leq i < j \leq n, |u_i - u_j| < h\}$, with the number of proximity model pairs given by $\mathcal{N}(h) = |S(h)|$. Under ideal conditions, the battle count $c_{ij}$ between models $i$ and $j$ is:
\begin{equation}
    c_{ij}(h) = \begin{cases} \mathcal{C}/\mathcal{N}(h) & \text{if } (i,j) \in S(h) \\ 0 & \text{otherwise} \end{cases} 
\end{equation}
where $\mathcal{C}$ is the total battle count, $\mathcal{N}(h)$ is a piecewise constant function (step function).  As $h$ increases, the number of \textbf{ proximity model pairs grows at each discontinuity point} $h_{(t)}$, when $h$ becomes sufficiently large, proximity sampling degenerates to random sampling. 

In Section \ref{Discussion:theoretical}, we measure the variance of the MLE using the trace of the pseudo-inverse of the \textbf{weighted Laplacian matrix}, denoted as $\varphi(h) = \text{tr}[L(w(h))^\dagger]$. To further characterize the global behavior of $\varphi$, we quantify its local variation by evaluating the change $\Delta \varphi$ at discontinuity points.

Let $S_t^{-} = S(h_{(t)} - \epsilon)$, $N_t^{-} = |S_t^{-}|$, $S_t^{+} = S(h_{(t)} + \epsilon)$. When $h$ transitions from $h_{(t)} - \epsilon$ to $h_{(t)} + \epsilon$ ($\epsilon \to 0^+$) at each discontinuity point, new proximity model pairs are introduced. 
Let $\Delta S_t = \{(i,j) \mid |u_i - u_j| = h_{(t)}\}$ denote these additional pairs, then $S_t^{+} = S_t^{-} \cup \Delta S_t$, $N_t^+ = |S_t^+|$. For each new model pair $(i,j) \in \Delta S_t$, the weight increment in \textbf{"ideal" setting} is given by: 
\begin{equation}
    \Delta w_{ij} = w_{ij}(h_{(t)} + \epsilon) - 0 = \frac{\mathcal{C} f_{ij}}{N_t^-},
\end{equation}
and for old model pair $(i,j) \in S_t^{-}$, 
\begin{equation}
    \Delta w_{ij} = \frac{\mathcal{C} f_{ij}}{N_t^+} - \frac{\mathcal{C}f_{ij}}{N_t^-} = -\mathcal{C} f_{ij}\frac{m_t}{N_t^+ N_t^-}. 
\end{equation}
Since the objective function $\varphi(w) = \mathrm{tr}[L(w)^{-1}]$ is continuous with respect to $w$, we can estimate its variation through Taylor expansion:
\begin{equation}
   \Delta \varphi(w) \approx \sum_{(i,j) \in S_t^+} \frac{\partial \varphi}{\partial w_{ij}}\bigg|_{L_t^{\dagger}} \Delta w_{ij}, 
\end{equation}
where $L_t^{\dagger} = L(w(h_{t} - \epsilon))^{\dagger}$, partial derivative  $\frac{\partial \varphi}{\partial w_{ij}} = -\text{tr}(L_t^{\dagger} A_{ij} L_t^{\dagger}) = -(e_i - e_j)^T (L_t^{\dagger})^2 (e_i - e_j) < 0$. 

Substituting $\Delta w_{ij}$ into the above expression, we obtain:
\begin{equation}\label{equ:delta_phi}
    \Delta \varphi \approx -\underbrace{\sum_{(i,j) \in \Delta S_t} \left|\frac{\partial \varphi}{\partial w_{ij}}\bigg|_{L_t^{\dagger}}\right| \cdot \frac{\mathcal{C} f_{ij}}{N_t^+}}_{A_t: \text{benefit from new pairs}} + \underbrace{\sum_{(i,j) \in S_t^-} \left|\frac{\partial \varphi}{\partial w_{ij}}\bigg|_{L_t^{\dagger}}\right| \cdot\frac{\mathcal{C} f_{ij} m_t}{N_t^- N_t^+}}_{B_t: \text{cost of budget reallocation}}.
\end{equation}
The derived expression for \(\Delta \varphi\) captures the \textbf{trade-off between two competing effects} when expanding the proximity sampling width \(h\): (1) the \textbf{gain} from adding new edges (\(A_t\)), which improves connectivity and reduces variance, and (2) the \textbf{cost} of budget redistribution (\(B_t\)), which dilutes the comparison density on existing edges. Critically, the sign of \(\Delta \varphi\) determines whether expanding \(h\) is beneficial: if \(A_t > B_t\), the net reduction in total variance justifies including additional pairs (edges).

\begin{figure}[ht]
    \centering
    \includegraphics[width=0.75\linewidth]{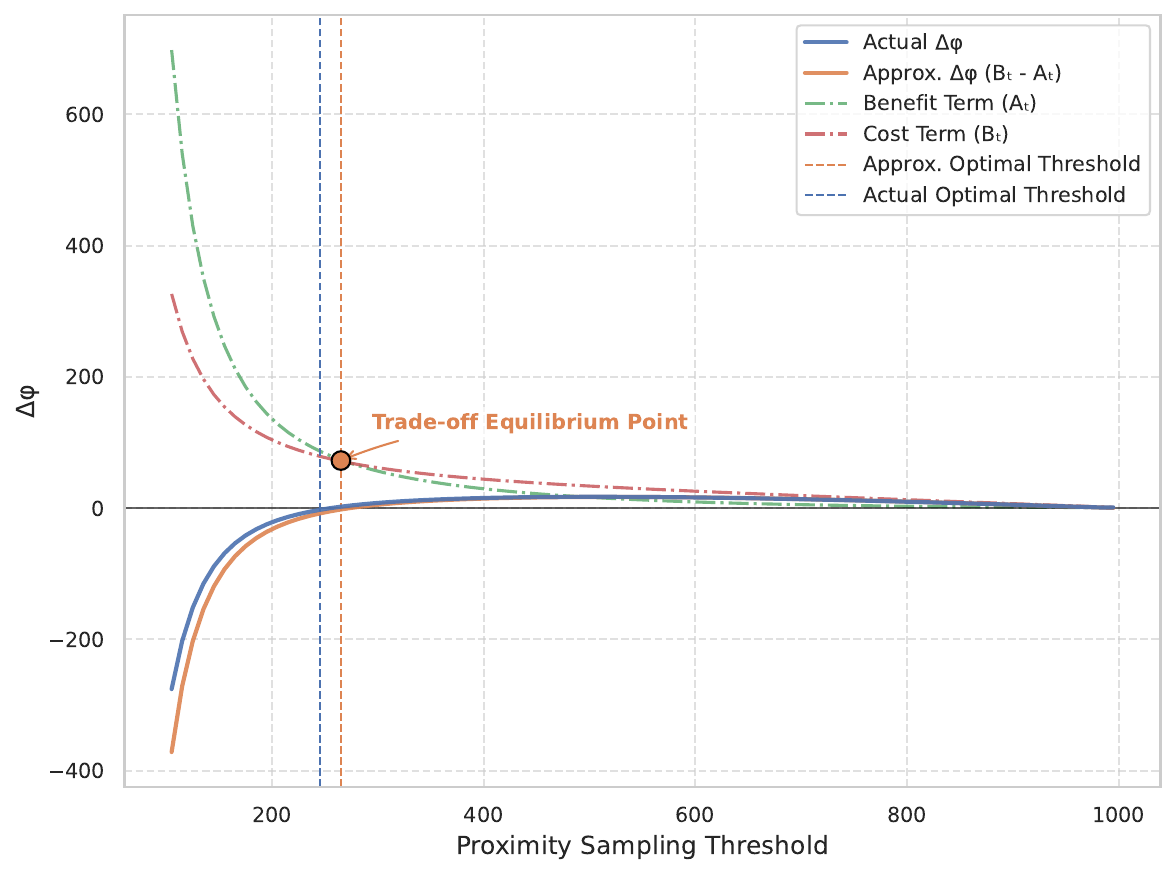}
    \caption{Decomposition of the $\Delta\varphi$ with Proximity Threshold. This figure illustrates the actual change (Actual $\Delta\varphi$), its first-order theoretical approximation (Approx. $\Delta\varphi$), and the two components of the approximation: the Benefit Term ($A_t$) from adding new model pairs and the Cost Term ($B_t$) from budget reallocation. The intersection of $A_t$ and $B_t$ marks the \textbf{trade-off equilibrium point}, which accurately predicts the optimal sampling threshold where the total variance is minimized.
}\label{fig:delta_phi_decomposition}
\end{figure}

To better understand why $\varphi(h)$ exhibits a U-shaped curve, we estimate the variation of the two components $A_t$ and $B_t$ of $\Delta\varphi(h)$ under an "ideal" setting in Figure \ref{fig:delta_phi_decomposition}. We observe that both \( A_t \) and \( B_t \) decrease as \( h \) increases. In the initial stage, \( A_t > B_t \), indicating that the benefit of introducing new proximity model pairs outweighs the cost. As \( h \) grows, \( B_t \) eventually surpasses \( A_t \). The optimal \( h^* \) is achieved when \( A_t = B_t \), i.e., when cost and benefit reach equilibrium.  Intuitively, when \( h \) is small, increasing it enhances the overall connectivity and transitivity of the comparison graph. However, when \( h \) becomes too large, some comparison efforts are wasted on less informative battles. Furthermore, Figure \ref{fig:delta_phi_decomposition} also demonstrates the approximation error introduced by Taylor expansion. While there exists some discrepancy between the true $\Delta\varphi$ and the approx $\Delta\varphi$ given by Equ \ref{equ:delta_phi}, leading to minor deviations in the obtained optimal $h^*$, the overall conclusions remain consistent.


\section{Simulation on the Chatbot Arena Dataset}\label{sec:simu_chatbot_arena}
To empirically validate the effectiveness of our proposed sampling strategy in a realistic setting, we conduct a simulation experiment using the open-source pairwise comparison data from Chatbot Arena. After excluding tied results, the dataset comprises 1,093,875 comparisons across 129 unique models. We then calculate the reference Elo scores (Arena scores) of each model using maximum likelihood estimation (MLE), which serve as reliable representations of relative capabilities due to the large scale of the dataset.
The simulation protocol is designed to mimic the aforementioned ranking process, starting with a cold-start phase, using the initial 20\% of the data (218,776 records) to establish preliminary Elo scores. Subsequently, the remaining data is introduced progressively in chronological order. In each round of simulation, newly introduced models are assigned an initial score through \textbf{placement matches}, with the number of comparison matches set to $T=10$. For existing models, we apply two competing strategies to select opponents for comparison:
\begin{itemize}
  \item \textbf{Proximity Sampling:} Our proposed method, governed by the proximity threshold $h\in(50,1000)$ and the minimum comparison count $m\in\{0,1,3\}$.
  \item \textbf{Uniform Sampling:} A baseline strategy sampling pairs from the Chatbot Arena dataset chronologically. It does not perform theoretical uniform sampling, but rather replicates the empirical data distribution resulting from Chatbot Arena's native sampling mechanism.
\end{itemize}
The Elo scores for all models are recalculated every three simulated days using all accumulated data, and this process continues until the entire dataset has been processed.

\begin{figure}[ht]
    \centering
    \begin{subfigure}[b]{0.49\textwidth}
        \includegraphics[width=\textwidth]{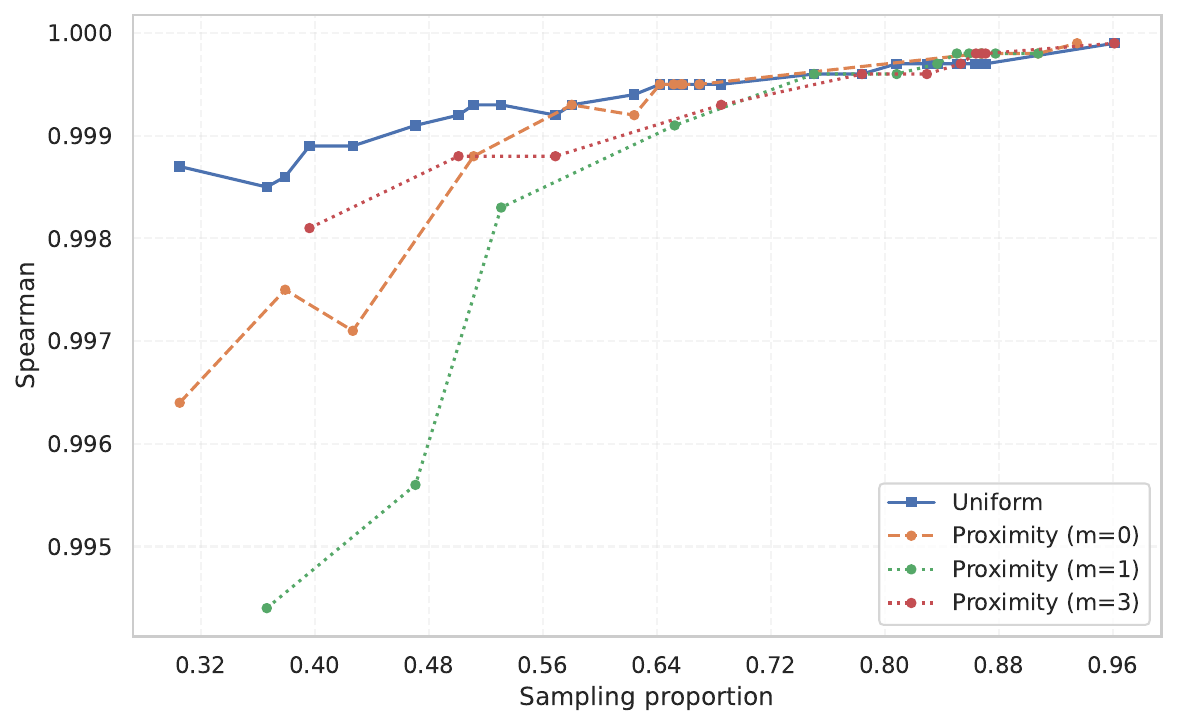}
        \label{fig:spearman}
    \end{subfigure}
    \begin{subfigure}[b]{0.49\textwidth}
        \includegraphics[width=\textwidth]{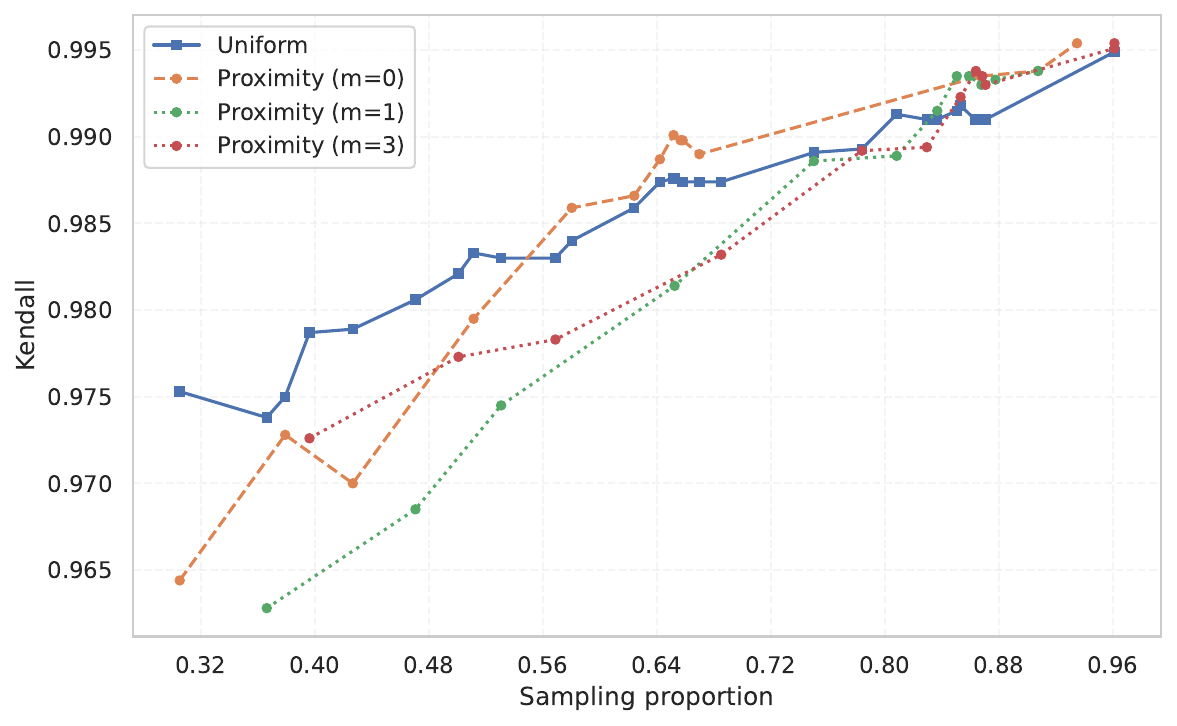}
        \label{fig:kendall}
    \end{subfigure}
    \begin{subfigure}[b]{0.49\textwidth}
        \includegraphics[width=\textwidth]{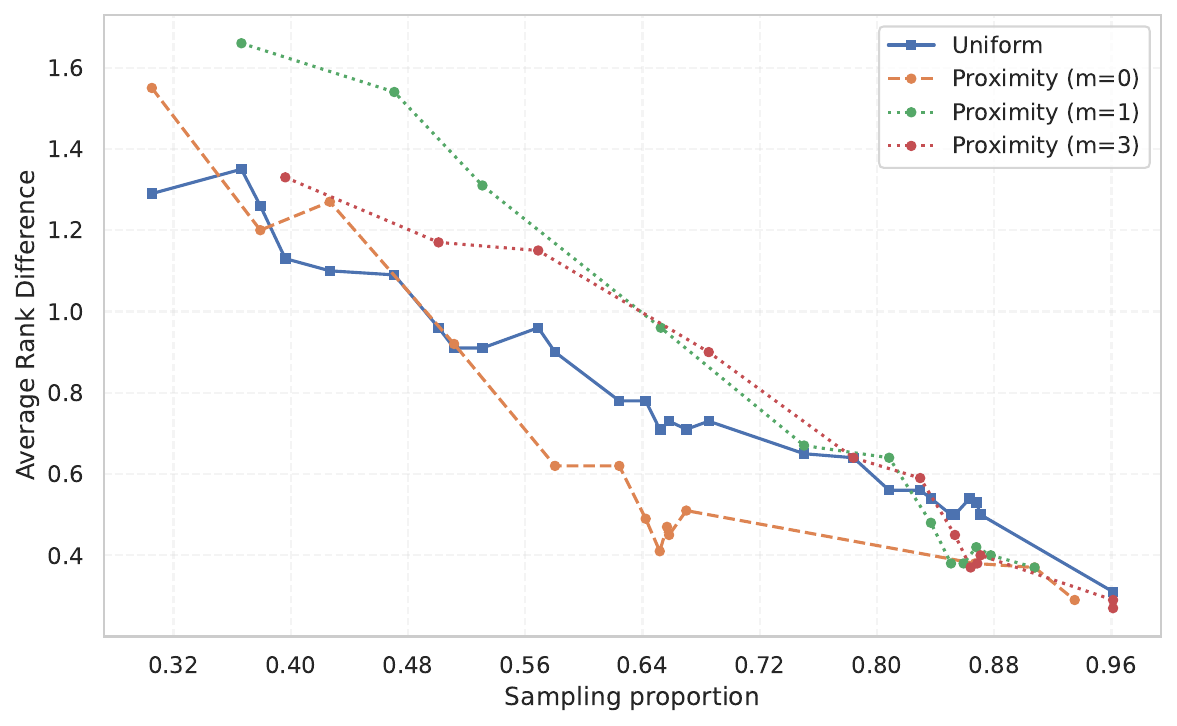}
        \label{fig:avg_rank_acc}
    \end{subfigure}
    \begin{subfigure}[b]{0.49\textwidth}
        \includegraphics[width=\textwidth]{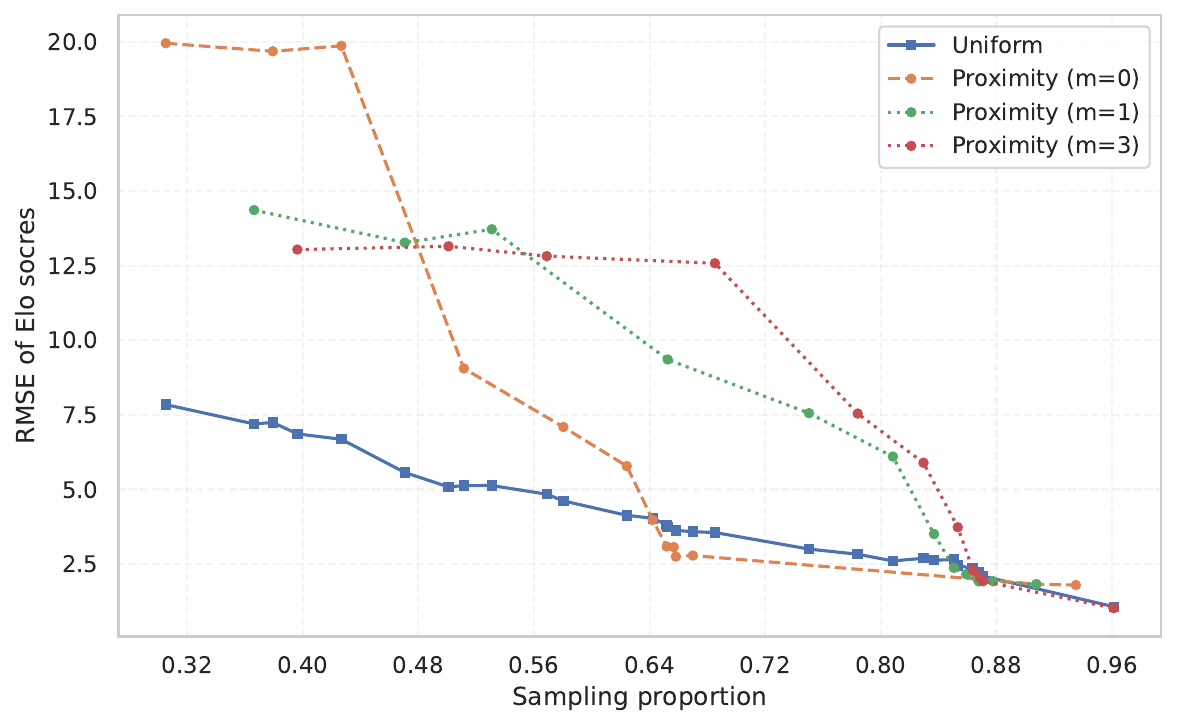}
        \label{fig:elo_emse}
    \end{subfigure}
    \caption{Simulation results under different sampling strategies. The parameter $m$ represents the minimum number of comparisons required with other selected models. The Sampling proportion reflects the magnitude of $h$.}
    \label{fig:simu_chatbot}
\end{figure}

We assess the accuracy of the resulting rankings using Spearman's Rank Correlation ($\rho$), Kendall's Rank Correlation ($\tau$), and the Average Rank Difference which calculates the mean absolute difference between each model's predicted and ground-truth rank. To evaluate the precision of the Elo scores themselves, we use the root mean squared error (RMSE) between the predicted and the reference Elo scores. Figure \ref{fig:simu_chatbot} presents a comparative analysis of these strategies, illustrating how different sampling methods affect ranking stability and score accuracy, demonstrating the effectiveness and advantages of our proximity sampling strategy in building a more reliable leaderboard.

\textbf{The Trade-off of the Neighborhood Threshold. }
A significant finding is that the efficacy of the Proximity Sampling strategy depends on the proximity threshold $h$. As shown in the results for the \textit{ Proximity (m=0)} strategy, our approach does not uniformly outperform the baseline, particularly in the early stages when $h$ is small. The turning point occurs when the $h$ increases to a relatively large value—empirically observed at $h \approx 250$ in the simulation, where the proportion of sampled comparisons exceeds 0.56, the performance metrics of our strategy begin to match or surpass those of the uniform sampling baseline.

This phenomenon reveals a critical trade-off between information utility and comparison connectivity. Theoretically, a small $h$ maximizes the information obtained from each comparison by focusing on discerning subtle differences between models of similar capabilities. However, in practice, an overly restrictive $h$ results in a globally sparse comparison connectivity, where many models lack sufficient comparison paths to others. This sparsity can lead to significant errors in the estimated Elo scores using MLE, thereby undermining the advantages gained from higher information utility. As $h$ increases, the comparison connectivity is enhanced, which stabilizes the global ranking estimation. At this point, the superior information utility of Proximity Sampling becomes evident, leading to improved performance in both ranking accuracy and Elo scores precision over the uniform sampling baseline.

\textbf{Data Efficiency. }
The simulation provides strong evidence of the superior data efficiency of the Proximity Sampling strategy. The results indicate that our approach can achieve the desired level of ranking accuracy using significantly fewer comparisons than uniform sampling. For example, the proximity sampling strategy ($m=0$) achieves an RMSE of Elo scores of 3.07, utilizing approximately 65.67\% of the comparisons. In contrast, the uniform sampling baseline requires more than 75.03\% of the comparisons to reach a comparable result of 3.00. By focusing resources on the most informative comparisons, those with high outcome uncertainty, our strategy enables the rapid and cost-effective construction of reliable model evaluation leaderboards.

\textbf{The Effect of the Minimum Comparison Constraint. }
An additional finding from our simulation relates to the role of the minimum number of comparisons $m$. The results presented in Figure \ref{fig:simu_chatbot} indicate that for a fixed sampling proportion, strategies with Proximity ($m>0$) perform worse than the Proximity ($m=0$) strategy. Furthermore, their performance degrades as $m$ increases, and in many cases, they are even inferior to the uniform sampling baseline.  
As established in Section 3.2, the foundational premise of Proximity Sampling is to maximize information gain by focusing comparisons within a trust region threshold $h$ where outcome uncertainty is highest. In essence, the Proximity ($m=0$) strategy adheres strictly to this principle, and the Proximity ($m>0$) strategy is intended to enforce a minimum level of comparison connectivity for each model. However, due to the inherent sparsity of the real-world Chatbot Arena dataset, finding sufficient neighbors within the narrow proximity region is often not possible. To satisfy the $m$ requirement, the Proximity ($m>0$) strategies allocate a portion of their finite comparison budget to matches that are less informative and have more deterministic outcomes. This forced supplementation leads to a systematic dilution of information quality, causing a slower and more imprecise overall MLE convergence of the Elo scores.

\section{Bootstrap Analysis of Elo Estimation Stability}
\label{sec:bootstrap_analysis}


\begin{figure}[ht]
    \centering
    \begin{subfigure}{0.95\textwidth} 
        \includegraphics[width=\textwidth]{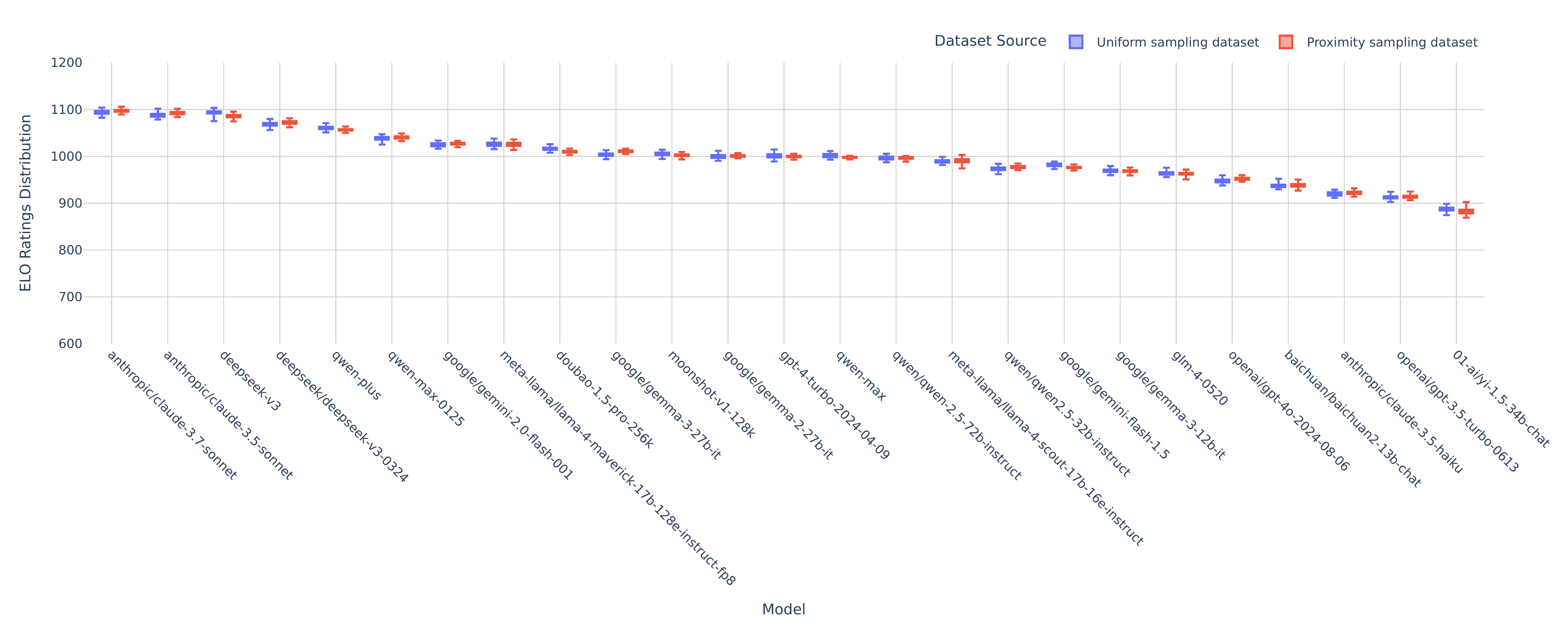}
        \caption*{(a) Distribution of Bootstrap Elo Ratings}
        \label{fig:combined_bootstrap}
    \end{subfigure}
    \vspace{0.01\textwidth}
    \begin{subfigure}{0.99\textwidth} 
        \includegraphics[width=\textwidth]{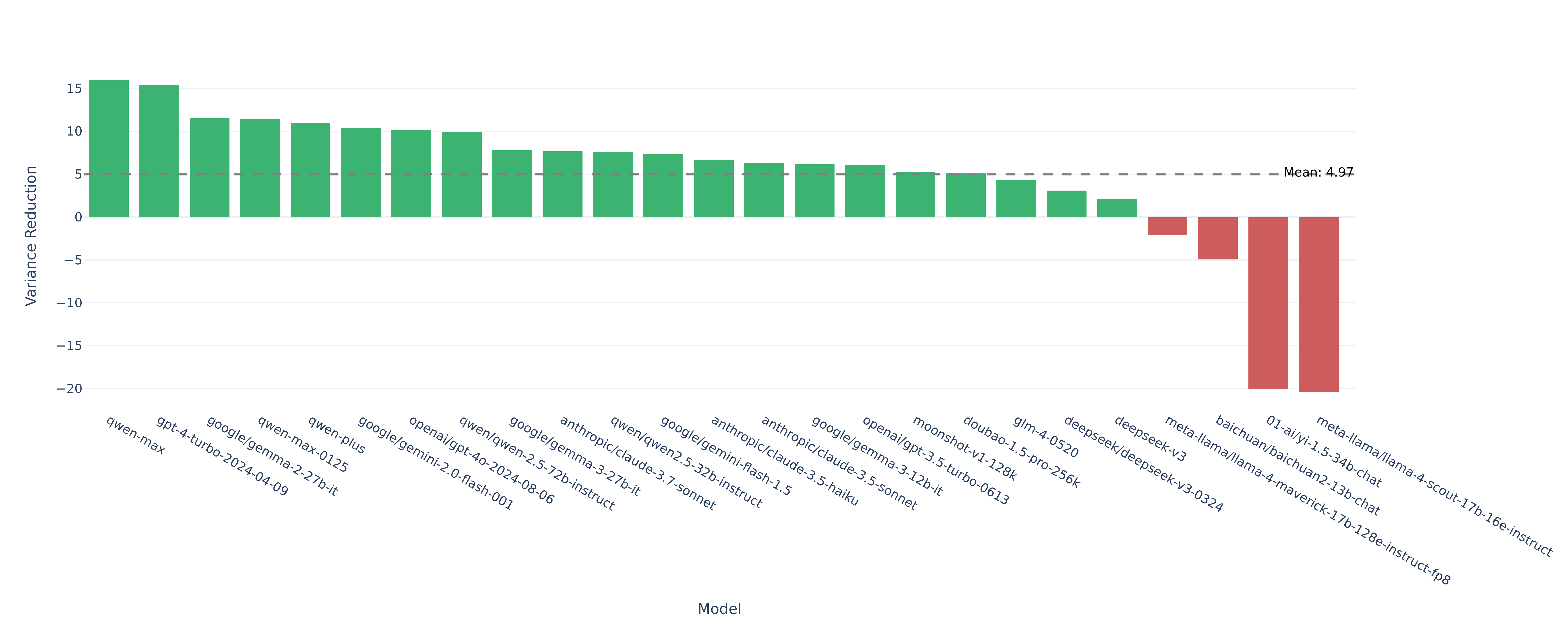}
        \caption*{(b) Variance Reduction in Elo Estimates}
        \label{fig:variance_difference}
    \end{subfigure}
    \begin{subfigure}{0.95\textwidth} 
        \includegraphics[width=\textwidth]{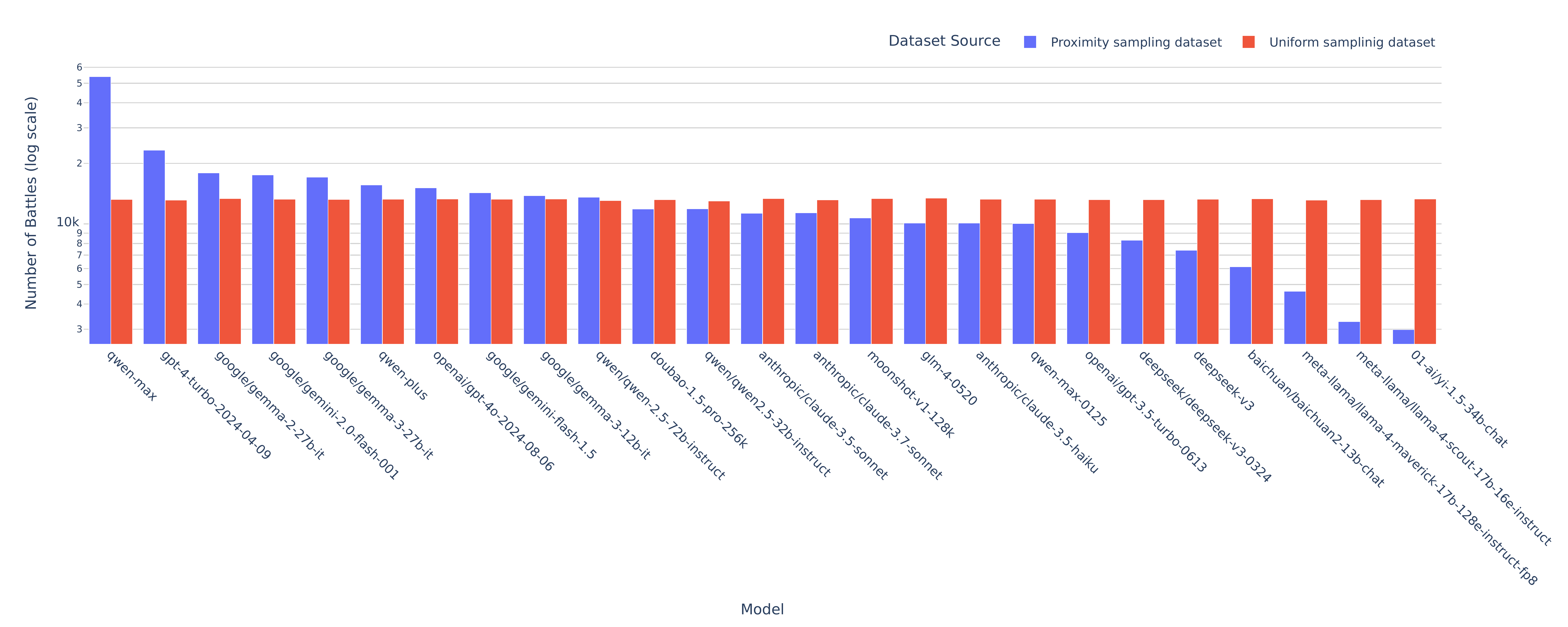}
        \caption*{(c) Model Comparison Count Distribution}
        \label{fig:battle_distribution}
    \end{subfigure}
    \caption{Stability Analysis of Elo Rating Estimation via Bootstrap Resampling. In (b), positive values indicate more stable Elo estimates under proximity sampling, while negative values suggest better stability with uniform sampling. (c) shows the logarithmic distribution of battles across models.}
    \label{fig:bootstrap}
\end{figure}

Although our primary simulation demonstrates the data efficiency of Proximity Sampling in achieving accurate rankings, a crucial question is the statistical reliability of the calculated Elo scores. A reliable ranking should not only be accurate on average, but also stable and less susceptible to random fluctuations in the sampled data. To investigate this, we conduct a bootstrap analysis to compare the stability of Elo ratings derived from two distinct data distributions: 
\begin{itemize}[leftmargin=2em]
   \item \textbf{Proximity Sampling Dataset:} The full dataset accumulated after implementing our Proximity Sampling algorithm, encompassing all data collected to date. To avoid potential bias from insufficient samples and ensure a meaningful comparison, we only retain model pairs with more than $2,000$ pairwise, 
    comparisons.
   \item \textbf{Uniform Sampling dataset:} A synthetic dataset generated to provide a controlled baseline for comparison. It is constructed with \textbf{the same set of models, Elo ratings, and total number of battles} as the Proximity Sampling Dataset, but the battle pairs are selected uniformly at random.
\end{itemize}
The goal is to compare the variance of Elo ratings estimates generated from datasets of equal size, collected through Proximity Sampling versus Uniform Sampling, to determine which strategy yields more statistically robust results. The analysis employs 100 rounds of bootstrap, with each round drawing a new dataset of the same original size via sampling with replacement.

Figure \ref{fig:bootstrap} demonstrates the effects of Uniform and Proximity Sampling strategies on Elo ratings stability.  As shown in Figure \ref{fig:bootstrap} (b), the Elo variance for \textit{qwen-max} is reduced by $15.9$, and for \textit{gpt-4-turbo-2024-04-09} is reduced by $15.4$, indicating a much more stable and reliable ratings under Proximity Sampling. Crucially, this stability gain is not merely an artifact of data volume. Figure \ref{fig:bootstrap} (c) reveals that among the $15$ models that received fewer battle counts under Proximity Sampling, $11$ still exhibited lower variance, indicating more stable Elo estimates. 

However, we do observe negative variance reduction for a few models, such as \textit{01-ai/yi-1.5-34b-chat} and \textit{meta-llama/llama-4-scout-17b-16e-instruct}. We attribute this phenomenon to their significantly lower battle counts, as they were recently added to the platform. Models with such sparse data are inherently more susceptible to statistical fluctuations during the bootstrap resampling process, leading to higher variance in their Elo estimates. Although a small number of models exhibit a negative variance reduction, the overall trend with a mean-variance reduction of $4.97$ across all qualifying models supports the superiority of our method in generating statistically robust rankings.

\end{document}